\documentclass[]{jingdong}
\usepackage[toc,page,header]{appendix}
\usepackage{minitoc}
\usepackage{url}
\usepackage{amssymb}
\usepackage[utf8]{inputenc}
\usepackage{microtype}
\usepackage{booktabs}
\usepackage{pifont} 
\usepackage{multirow}
\usepackage{makecell}
\usepackage{paralist}
\usepackage{xspace}
\usepackage{color}
\usepackage{xcolor}
\usepackage{colortbl}
\usepackage{adjustbox}
\usepackage{hyperref} 
\usepackage[edges]{forest}
\usepackage{tikz} 
\usepackage{caption}
\usepackage{amsfonts}
\usepackage[toc,page,header]{appendix}
\usepackage[utf8]{inputenc}
\usepackage{bbm}
\usepackage{enumitem}
\usepackage{pgfplots}
\pgfplotsset{compat=1.18}

\usepackage{subcaption}   % for subtable
\usepackage{pifont}

\usepackage{threeparttable}  % for g1 table
\usepackage{graphicx}  % same as above

\definecolor{seedc}{RGB}{7, 92, 173}

\newcommand{\name}[1]{GM-0}

\newcommand{\hardware}[1]{ByteMini}

\renewcommand{\paragraph}[1]{\vspace{0.1em}\noindent\textbf{#1}}

\usepackage{tabularx}       % 自动列宽/固定列宽
\usepackage{latexsym}
\usepackage[T1]{fontenc}
\usepackage[utf8]{inputenc}
\usepackage{microtype}
\usepackage{inconsolata}
\usepackage{graphicx}
\usepackage{hyperref}       % hyperlinks
\usepackage{url}            % simple URL typesetting
\usepackage{booktabs}       % professional-quality tables
\usepackage{amsfonts}       % blackboard math symbols
\usepackage{nicefrac}       % compact symbols for 1/2, etc.
\usepackage{stackengine}
\usepackage{microtype}      % microtypography
\usepackage{colortbl}
\usepackage{xcolor}
\usepackage{amsmath}
\usepackage{amssymb}
\usepackage{amsthm}
\usepackage{mathrsfs}
\usepackage{pifont}
\usepackage{MnSymbol}
\usepackage{balance}
\usepackage{enumitem}
\usepackage{listings}
\usepackage{xcolor}
\usepackage{natbib}
\usepackage{multicol}
\AtBeginDocument{%
  \providecommand\BibTeX{{%
    \normalfont B\kern-0.5em{\scshape i\kern-0.25em b}\kern-0.8em\TeX}}}

\makeatletter
\DeclareRobustCommand\onedot{\futurelet\@let@token\@onedot}
\def\@onedot{\ifx\@let@token.\else.\null\fi}

\usepackage{setspace}
\usepackage{mathtools}

\usepackage{multirow,booktabs}
\usepackage{subcaption}

\newcommand{\owo}[1]{\textsc{OAgents}}

\definecolor{lightgreen}{RGB}{144, 238, 144} 
\definecolor{lightred}{RGB}{255, 105, 97}

\newtcolorbox{promptbox}[2][Prompt]{
colback=black!5!white,
arc=5pt, 
boxrule=0.5pt,
fonttitle=\bfseries,
title=#1, 
before upper={\small}, fontupper=\fontfamily{ptm}\selectfont,
colframe=#2, % 使用传递的参数来设定 colframe
}
\definecolor{ogreen}{RGB}{34, 139, 34}

\usepackage{cleveref}
\theoremstyle{plain}

\theoremstyle{definition}

\theoremstyle{remark}
\usepackage{subcaption}

\usepackage[textsize=tiny]{todonotes}
\usepackage{fontawesome}
\usepackage{float}
\usepackage{graphicx}
\usepackage{subcaption}
\usepackage{booktabs}
\usepackage{wrapfig}

\title{JoyAI-RA 0.5: Scaling Robot Manipulation Learning via Dual Action Alignment}
\author{JoyAI-RA Team}

\affiliation{Joy Future Academy, JD}

\abstract{

Robot data is scarce, so generalist policies need to learn from heterogeneous sources, including human egocentric video, simulation, and real robots, which differ in supervision and embodiment, with action labels missing or mutually incompatible. Human egocentric data scale best but sit farthest from robot data, and naive pooling causes negative transfer rather than knowledge sharing. We propose JoyAI-RA 0.5, a generalist Vision-Language-World-Action (VLWA) framework that couples physical world-dynamics priors with visual semantics and scales manipulation learning across such data via dual action alignment. Implicit action alignment infers latent actions from visual transitions, enabling action-free human, simulation, and robot data to guide a latent-action-conditioned world model in learning physical dynamics. Explicit alignment grounds reliable human and robot trajectories in a unified physical action space through a canonical action representation and camera-frame chunk-relative end-effector actions. An inner-outer-loop reinforcement stage then pairs efficient task adaptation with foundation-policy improvement. On a real-world AgiBot benchmark, JoyAI-RA performs strongly on both seen tasks and unseen variations. The task score improves consistently as the volume of human egocentric pretraining data increases and shows no sign of plateauing at our largest scale. This suggests that abundant but weakly labeled human experience can be converted into a transferable training signal, making human video not merely a weak auxiliary source but a primary axis along which manipulation capability can be scaled.
}

\checkdata[Project Page]{\url{https://joyai-ra-05.github.io/}}

\begin{document}
\maketitle

% 天乐
\section{Introduction}
\label{sec:introduction}

\begin{figure}[t]
\centering
\includegraphics[width=\linewidth]{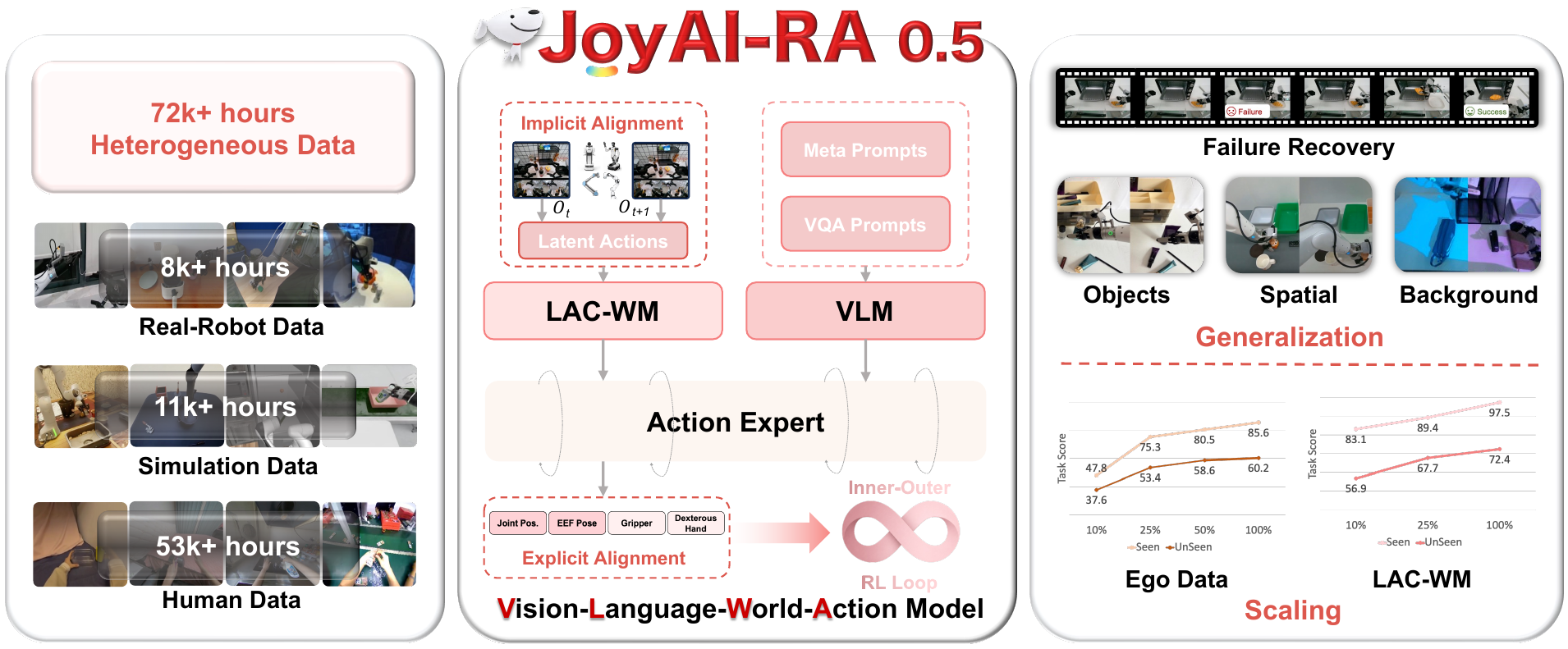}
\caption{\textbf{JoyAI-RA 0.5 is a Vision-Language-World-Action framework for scaling learning robot manipulation from heterogeneous data through dual action alignment.}}
\label{fig:teaser}
\vspace{-10pt}
\end{figure}

Robot manipulation data are scarce and expensive to acquire, limiting the scalability of generalist manipulation policies trained solely on robot demonstrations. A promising alternative is to leverage heterogeneous data sources, including human egocentric videos, simulation, and robot demonstrations, which together offer orders of magnitude more diverse manipulation experience. Among them, human egocentric video is particularly attractive because of its scale and diversity, yet also the most difficult to exploit for robot learning: most videos lack reliable action annotations, and even recovered hand trajectories differ fundamentally from robot control commands. The central challenge is therefore to derive shared, transferable supervision from heterogeneous data, rather than naively pooling incompatible sources.

% Robot manipulation data are scarce and costly to acquire. Scaling generalist policies therefore requires heterogeneous sources—human egocentric video, simulation, and real robots—that differ in supervision and embodiment, with action labels often missing or incompatible. Human egocentric video is the most scalable and diverse of these, yet sits farthest from robot data: most clips lack reliable action labels, and even recovered hand trajectories differ from robot commands and are measured in a moving head-mounted camera frame. The central obstacle is thus how to unify these sources into a shared, transferable form of supervision—naively pooling them induces negative transfer rather than knowledge sharing.

Existing Vision-Language-Action (VLA) models inherit rich semantic priors from pretrained vision-language models (VLMs)~\cite{kim2024openvla,bjorck2025gr00t,black2024pi_0,intelligence2025pi_, zhang2026joyai}, enabling strong instruction following and semantic reasoning. However, they lack an understanding of physical dynamics and therefore cannot effectively leverage the abundant unlabeled human videos that capture how the physical world evolves under manipulation. World-action models (WAMs) learn predictive physical dynamics directly from pixels~\cite{chen2025moto,hu2024videopredictionpolicy,kim2026cosmos}, but lack semantic grounding and generally rely on action annotations unavailable in human videos. Consequently, VLA models capture semantics without dynamics, whereas WAMs capture dynamics without semantics, limiting the benefits of heterogeneous data.

To address these challenges, we propose \textbf{JoyAI-RA 0.5}, a Vision-Language-World-Action (VLWA) framework that integrates physical world-dynamics priors with visual semantics and enables scalable manipulation learning from heterogeneous data through dual action alignment, as shown in Figure~\ref{fig:teaser}. The framework integrates a Vision-Language Model (VLM), a Latent-Action-Conditioned World Model (LAC-WM), and a Flow-Matching Action Expert to jointly represent task semantics, predict physical evolution, and generate executable actions. The key idea is to align heterogeneous data through two complementary supervision channels: implicit alignment and explicit alignment. Implicit action alignment exploits object-dynamics regularities shared across embodiments by using latent actions inferred from visual state transitions as conditioning variables for the world model, thereby decomposing multimodal future prediction into simpler transition-specific dynamics modes and converting large-scale action-free human videos into scalable, transferable dynamics supervision. Explicit action alignment leverages action-annotated data by mapping human and robot trajectories into a canonical chunk-relative action representation, providing physically grounded supervision across embodiments. Together, the two alignment mechanisms turn action-free and action-annotated data into complementary learning signals, enabling scalable policy learning across diverse data sources and embodiments. To further improve policy adaptation during deployment, JoyAI-RA incorporates an inner-outer loop reinforcement learning strategy that combines rapid online residual adaptation with periodic foundation policy updates, continuously improving both task performance and generalization.

We evaluate JoyAI-RA on a real-world AgiBot G1 robot platform across pick-and-place, high-precision manipulation, and long-horizon tasks. JoyAI-RA achieves strong performance on seen tasks while generalizing effectively to novel objects, spatial configurations, backgrounds, and lighting conditions. 
% Ablation studies show that both channels of dual alignment, together with the introduced world model, contribute substantially to downstream performance, particularly under unseen conditions.
Most importantly, JoyAI-RA exhibits a clear scaling trend with human egocentric pretraining data, with performance continuing to improve at our largest training scale, suggesting substantial headroom for further scaling. These results demonstrate that dual action alignment transforms abundant but weakly labeled human experience into transferable supervision, establishing human egocentric video not merely as auxiliary data, but as a primary driver for scaling robot manipulation.

\section{Related Work}
\label{sec:related_work}
% xiangkai
\paragraph{World Models for Robotics.}
Generative world models~\citep{hafner2023dreamer3} have recently emerged as a promising paradigm for embodied intelligence, enabling agents to predict future physical states for planning and control. 
% Early model-based reinforcement learning methods, such as Dreamer series~\citep{hafner2019dreamer1,hafner2019dreamer2,hafner2023dreamer3}, learn compact predictive states for downstream control. 
More recent embodied world models leverage generative video and world-action modeling objectives to learn robot-environment dynamics from large-scale heterogeneous data~\citep{du2024videolanguageplanning,kim2026cosmos}. 
Existing video-based robot world models generally follow two lines: the imagine-then-act paradigm, which predicts future visual states before recovering actions via inverse dynamics, often introducing latency in closed-loop deployment~\citep{ko2024learningtoact,pai2025mimic,chen2025moto,hu2024videopredictionpolicy,tian2025predictiveinverse,yuan2026fastwam}; 
and joint world-action modeling, which unifies future visual dynamics and actions within a shared generative architecture~\citep{li2026causal,ye2026worldzeroshot,yuan2026fastwam,ye2026gigaworld}. 
A critical evolution in recent world modeling is the shift away from pixel-level reconstruction toward latent feature forecasting. Pioneered by JEPAs~\citep{assran2025vjepa2}, these approaches distill underlying physical and temporal patterns directly from abstract representations. 
This latent-centric design has subsequently catalyzed diverse generative innovations, ranging from latent diffusion and action-conditioned structured forecasting in Motus and other works~\citep{bi2025motus,lyu2026lda,team2026motubrain} to unified denoising paradigms~\citep{li2026causal,guo2024predictionwithaction,zhu2025unifiedworldmodel,cen2025worldvla,ye2026worldzeroshot,zhang2026dreamvla} that seamlessly merge visual anticipation with motor generation. 
% To mitigate the computational overhead of these complex pipelines, architectural optimizations like Fast-WAM~\citep{yuan2026fastwam} achieve rapid deployment by entirely circumventing the video decoding step during inference. 
% Overall, these advances highlight that explicit future-state modeling enhances sample efficiency, robustness, and generalization.
Despite these advancements, purely visual or physics-driven world models still face fundamental limitations: they capture physical dynamics from pixels but lack native semantic grounding and language understanding, making it difficult to align with high-level task instructions. 
% Moreover, relying on explicit pixel-level generation or strictly annotated action spaces restricts their scalability across heterogeneous data (e.g., unlabeled human videos). 
To address these bottlenecks, JoyAI-RA 0.5 introduces a Latent-Action-Conditioned World Model (LAC-WM) that synergizes with a Vision-Language Model (VLM) under a complementary architecture. 
% Specifically, the LAC-WM focuses on predicting future physical states to capture dynamic priors, while the VLM handles sub-task semantics. 
% Crucially, we employ a Knowledge Injection (KI) mechanism to fuse the WM's physical priors directly into the Action Expert, ensuring that executable actions are grounded in both semantic intent and physical feasibility.

\paragraph{Vision-Language-Action Models.}
Large-scale Vision-Language-Action (VLA) pretraining has become the foundation for robot policies, evolving through two primary paradigms. 
Early systems cast control as autoregressive (AR) token generation, utilizing direct discretization~\citep{brohan2022rt,zitkovich2023rt,kim2024openvla,galaxea2026g05,pertsch2025pi0fast} or learning-based vector quantization~\citep{wang2025vqvla,ma2026unifying}. 
This approach scales poorly, as high-frequency control becomes prohibitively expensive due to the rapid growth of autoregressive tokens. 
This bottleneck pushed the field toward VLM-as-encoder architectures, where a pretrained VLM supplies hidden states or KV cache to a separately trained flow-matching or diffusion expert that predicts continuous action chunks~\citep{black2024pi_0,intelligence2025pi_,bjorck2025gr00t,li2024cogact,zhao2026sim2real}. 
To enhance long-horizon reasoning, recent works integrate chain-of-thought (CoT) mechanisms into VLM-as-actor frameworks. 
These approaches either employ modular ``bolt-on'' strategies that treat high-level planning as an external interface~\citep{bjorck2025gr00t,li2025hamster}, or embed CoT directly into the autoregressive stream to jointly predict intermediate reasoning steps and actions within a unified sequence~\citep{zawalski2024robotic,zhang2026atomicvlaunlockingpotentialatomic,zhao2025cotvla,sun2025emmax,intelligence2025pi_}. 
However, existing VLA models remain constrained by weak physical reasoning and limited scalability across heterogeneous data. 
To address these limitations, JoyAI-RA 0.5 introduces a complementary architecture and a novel data alignment strategy. 
% Specifically, we decouple semantic reasoning from physical dynamics, where the VLM excels at extracting task semantics and predicting coarse, low-frequency ``Fast Actions'', the WM captures underlying physical dynamics by predicting future states. 
% To overcome the scalability bottleneck caused by scarce and heterogeneous data, our novel data alignment strategy employs a dual-alignment paradigm that seamlessly integrates large-scale, unannotated human ego-view videos with multi-embodiment robot and simulation data.

\paragraph{Latent Action Learning.}
Existing VLAs and WAMs face two fundamental challenges: the inability to acquire primitive-level action representations for cross-embodiment alignment~\citep{wang2024scaling,chen2024mirage}, and a severe lack of scalable, diverse robotic data compared to human data~\citep{yuan2026qwenrobotmanip,wang2026humanego}. 
To address this dual dilemma, Latent Action Models (LAMs)~\citep{ye2025latent,bu2025univla} have emerged as a prevailing paradigm. 
Originating from video modeling~\citep{bruce2024genie}, LAMs resolve the alignment challenge by using visual supervision to map diverse action primitives into a shared, embodiment-agnostic latent manifold~\citep{bu2025univla,nikulin2025latentactionlearning,liang2025clam,bjorck2025gr00t}. 
Simultaneously, they overcome the data bottleneck via inverse dynamics self-supervised learning~\citep{chen2025moto}. 
By predicting latent actions from frame-to-frame state transitions, LAMs provide a promising route for leveraging unlabeled human videos to alleviate the scarcity of high-quality robotic data~\citep{liu2026lara}.
% However, existing LAMs predominantly rely on a single implicit alignment channel. 
% While effective for data scaling, the resulting latent actions are inherently not directly executable and often suffer from a representation gap when transferring to downstream executable control across diverse robot embodiments. 
% Furthermore, standalone LAMs lack a mechanism to explicitly bridge this latent scaling with embodiment-specific physical grounding, limiting their direct deployment in heterogeneous multi-robot settings. 
% To overcome these limitations, JoyAI-RA 0.5 introduces a novel Dual-Alignment Paradigm that fundamentally extends traditional LAMs.
Despite this promise, existing LAMs have rarely been scaled to substantially larger and more heterogeneous video corpora and typically infer latent actions from single-view visual transitions, limiting the interaction and geometric information captured in the latent space. JoyAI-RA 0.5 instead learns latent actions from multi-view transitions across large-scale human, simulation, and robot videos, producing a richer latent space that conditions the world model with transferable, action-sensitive dynamics priors.

\section{Data Preprocessing and Dual Alignment}\label{sec:data}

\subsection{Data Recipe}\label{sec:data_recipe}
\begin{figure}[!tbp]
\centering
\includegraphics[width=1.0\linewidth]{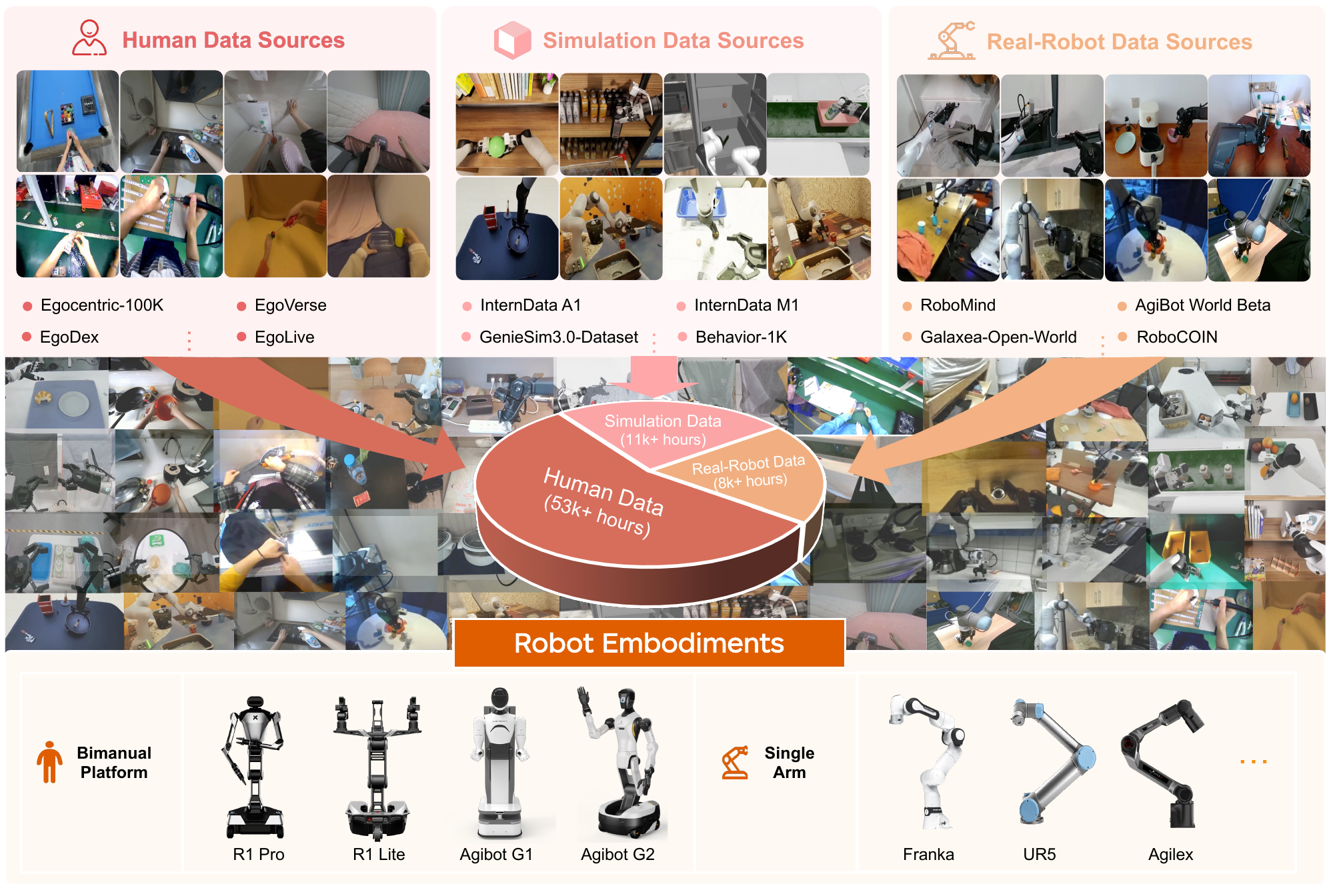}
\caption{
\textbf{Overview of the pre-training dataset.}
The pre-training dataset combines open-source Internet datasets, self-collected human and real-robot demonstrations, and simulation trajectories, totaling 53K+, 11K+, and 8K+ hours, respectively, across diverse bimanual and single-arm embodiments.
}
\label{fig:pre-training_dataset}
\end{figure}

As shown in Figure~\ref{fig:pre-training_dataset}, our pre-training corpus integrates three complementary data sources: human egocentric videos, simulation trajectories, and real-robot demonstrations. It comprises over 53K hours of human video, 11K hours of simulation data, and 8K hours of real-robot data, with the latter two spanning diverse bimanual and single-arm robot embodiments.

\paragraph{Egocentric Human Manipulation Data.}
% The human-data corpus combines our in-house EgoLive dataset~\citep{li2026egolive} with a collection of large-scale egocentric datasets, including Egocentric-100K~\citep{buildai2025egocentric100k}, Xperience-10M~\citep{ropedia2026xperience10m}, EgoVerse~\citep{punamiya2026egoverse}, Ego4D~\citep{grauman2022ego4d}, Ego-Exo4D~\citep{grauman2024egoexo4d}, EgoDex~\citep{hoque2025egodex}, and EPIC-KITCHENS-100~\citep{damen2022epickitchens100}. Collectively, these datasets capture diverse human-object interactions and provide rich visual-semantic supervision for learning task understanding, action decomposition, and manipulation behaviors. EgoLive\cite{li2026egolive} contains 20k+ hours of 60 FPS egocentric manipulation videos, covering 600+ real-world task scenarios across household, retail, and logistics environments. It provides rich annotations with 2,662 action categories, 53,400 object categories and 31,400 attribute categories, offering significantly high diversity. It encompasses both short-horizon interactions and long-horizon activities, providing valuable supervision for hierarchical planning and multi-step manipulation. Detailed statistics on the EgoLive dataset are provided in Sec.~\ref {subsec:egolive}.
The human-data corpus first aggregates a collection of large-scale egocentric datasets, including Egocentric-100K~\citep{buildai2025egocentric100k}, Xperience-10M~\citep{ropedia2026xperience10m}, EgoVerse~\citep{punamiya2026egoverse}, Ego4D~\citep{grauman2022ego4d}, Ego-Exo4D~\citep{grauman2024egoexo4d}, EgoDex~\citep{hoque2025egodex}, and EPIC-KITCHENS-100~\citep{damen2022epickitchens100}. These datasets provide broad visual and semantic coverage of human-object interactions across diverse environments and activities. We further augment this corpus with our in-house EgoLive dataset~\citep{li2026egolive}, which contains over 20K+ hours of egocentric manipulation videos captured at 60 FPS and spans more than 600 real-world task scenarios across household, retail, and logistics environments. EgoLive provides fine-grained annotations covering 2,662 action categories, 53,400 object categories, and 31,400 attribute categories, together with both short-horizon interactions and long-horizon activities. These annotations provide rich supervision for task understanding, action decomposition, hierarchical planning, and multi-step manipulation. Detailed statistics of EgoLive are provided in Sec.~\ref{subsec:egolive}.

% Unlike datasets centered on video-level descriptions, EgoLive provides operation-centric annotations, including high-level task descriptions and temporally grounded subtask labels. We additionally recover aligned hand poses and motion trajectories to provide fine-grained interaction supervision. Further details on the dataset’s scale and diversity are provided in Sec.~\ref{subsec:egolive}.

\paragraph{Simulation Data.}
We incorporate large-scale simulation trajectories from InternData-A1~\citep{tian2025interndataa1}, InternData-M1~\citep{internrobotics2025interndatam1}, Genie Sim 3.0~\citep{yin2026geniesim3}, and BEHAVIOR-1K~\citep{li2023behavior1k}. These datasets span diverse environments, objects, robot embodiments, and manipulation tasks, providing scalable supervision with explicit robot actions and physically grounded state transitions. Together, they help bridge human-centric visual-semantic learning and robot-centric action learning while improving generalization across tasks and embodiments.

\paragraph{Real-Robot Data.}
Our real-robot corpus integrates large-scale demonstrations from AgiBot World Beta~\citep{bu2025agibotworld}, Galaxea Open-World~\citep{jiang2025galaxeaopenworld}, RoboCOIN~\citep{wu2025robocoin}, RoboMIND~\citep{wu2024robomind}, RoboMIND 2.0~\citep{hou2025robomind2}, LET~\citep{leju2025let}, and the Baihu Dataset~\citep{openloong2026baihu}. These data cover diverse robot embodiments, tasks, objects, and real-world environments, while naturally capturing sensing noise, actuation errors, contact uncertainty, and hardware constraints. Such real-world supervision grounds learned knowledge in physical execution and improves policy robustness and transferability across robots and environments.

These heterogeneous sources are first processed through the unified annotation and curation pipeline described in Sec.~\ref{sec:data_processing}, and are then organized into compatible forms of supervision through the implicit and explicit alignment mechanisms introduced in Sec.~\ref{sec:dual_alignment}.

\subsection{Data Preprocessing 
Pipeline}\label{sec:data_processing}
\begin{figure}[!tbp]
    \centering
    \includegraphics[width=\linewidth]{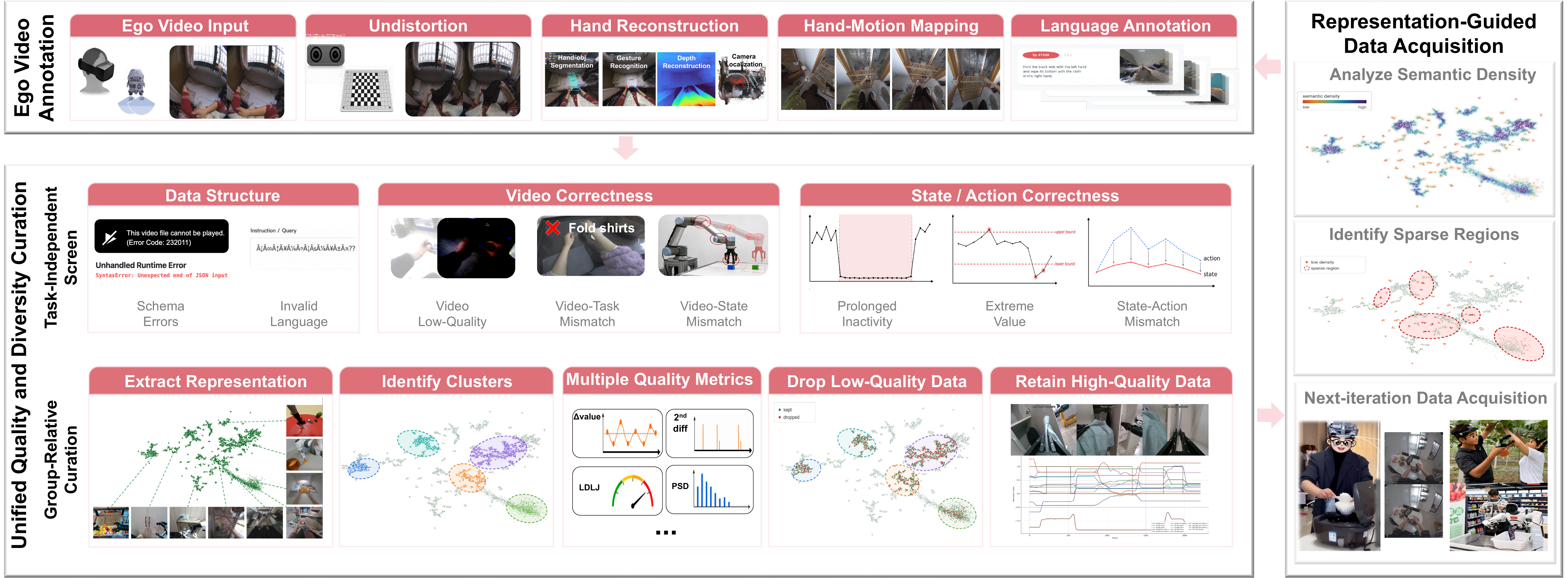}
    \caption{
    \textbf{Closed-Loop Data Preprocessing and Curation Pipeline.} Ego videos are first annotated automatically and then jointly processed with robot trajectories through unified quality and diversity curation, while representation-guided analysis directs subsequent data acquisition.
    }\label{fig:data_pipeline}
\end{figure}

The heterogeneous sources are unified within a closed-loop data curation pipeline. Raw human ego data are first processed to obtain manipulation-relevant visual and language annotations, together with estimated hand-motion trajectories. Reliable human trajectories are then mapped into a shared state-action representation compatible with robot data, allowing all sources to be curated under common validity, quality, and diversity criteria. The resulting corpus is further analyzed to identify underrepresented capabilities and guide subsequent data collection. Figure~\ref{fig:data_pipeline} illustrates the overall pipeline.

\paragraph{Egocentric Video Annotation and Hand-Pose Recovery.}
We process each raw egocentric video through a five-stage annotation pipeline. First, the input camera streams are synchronized, calibrated, undistorted, and rectified for stereo recordings such as EgoLive~\cite{li2026egolive}. We then reconstruct hand motion by combining hand-object segmentation, gesture recognition, depth reconstruction, and camera localization. Native trajectories from EgoDex~\cite{hoque2025egodex} are transformed into a consistent camera or world frame. For videos without native trajectories, a HaMeR-based estimator~\cite{pavlakos2024reconstructing} fits MANO~\cite{romero2022embodied} parameters to each view and refines them through stereo reprojection and temporal optimization, while ORB-SLAM3~\cite{campos2021orb} estimates camera motion and metric scale. The recovered bilateral wrist poses and fingertip positions are then mapped into the canonical hand-motion representation, together with confidence scores based on visibility, reprojection, and temporal consistency. Finally, the videos are segmented into atomic subtasks and annotated with structured language descriptions specifying the acting hand, manipulated object, and action.

Reliable hand trajectories with high confidence scores are used for explicit alignment in Sec.~\ref{sec:explicit}. Clips with valid visual and language content but unreliable or missing hand poses are still retained for latent-action-based LAC-WM pretraining in Sec.~\ref{sec:implicit}. This routing preserves useful ego-video supervision while preventing noisy physical trajectories from affecting explicit action learning.

\paragraph{Unified Quality and Diversity Curation.}
The collected episodes first undergo a \textit{task-independent validity screen} to remove invalid data. We verify data structure consistency by checking schema completeness, timestamp alignment, and annotation validity, removing episodes with missing fields, malformed records, or invalid language descriptions. Video streams are screened for visual quality issues, including black, corrupted, or blurred frames, as well as prolonged static segments. We further evaluate cross-modal consistency by comparing language instructions with observed manipulation behaviors and measuring the agreement between visual observations and reconstructed hand or robot states. For robot data, state-action correctness is assessed through trajectory analysis, including detection of prolonged inactivity, extreme values, and temporal mismatches between commanded actions and resulting state transitions. These checks ensure that only temporally aligned, physically plausible, and semantically consistent episodes proceed to subsequent curation.

The valid episodes are then processed through \textit{group-relative quality and diversity curation}. Since demonstration characteristics vary across tasks, embodiments, and control interfaces, episodes are compared only within compatible groups. We first extract compact episode representations using an embodied video encoder initialized from V-JEPA~2~\citep{assran2025vjepa2}, continued-pretrained on unlabeled ego and robot videos, and augmented with a Q-Former following Cosmos-Embed~\citep{nvidia2025cosmosembed1}. These representations capture scene, object, and manipulation semantics for measuring episode similarity. SemDeDup~\citep{abbas2023semdedup} is then applied within each task group to identify clusters of similar demonstrations. Within each cluster, we evaluate demonstrations using multiple complementary quality metrics, including derivative statistics, log dimensionless jerk, power spectral density, spectral arc length, motion spikes, and temporal entropy~\cite{sojib2026efficient,kulkarni2026learning}. Since representation similarity only indicates redundancy rather than quality, we retain high-quality representatives while preserving additional samples that provide diverse execution strategies or rare scene configurations. Low-quality and redundant episodes are removed based on their relative quality within each cluster, rather than fixed global thresholds. Finally, episode-level filtering results and invalid timestep indices are consolidated before chunk construction, and retained hand and robot trajectories are mapped into the unified physical representation introduced in Sec.~\ref{sec:explicit}. This produces a compact yet diverse corpus with reliable supervision across tasks and embodiments.

\paragraph{Representation-Guided Data Acquisition.}
The curated corpus is periodically analyzed in the learned embodied representation space to assess semantic coverage and representation density, with downstream performance providing an additional measure of corpus effectiveness. Sparse regions reveal underrepresented scene-object-behavior combinations and missing capabilities, which guide subsequent data acquisition. We adopt a hybrid collection strategy that combines open-ended human recording to capture diverse and unexpected interactions with targeted instructions to enrich specific underrepresented or deployment-relevant skills. Ego data primarily expand visual-semantic coverage, long-horizon behaviors, and general interaction patterns, while robot data provide embodiment-specific execution supervision. Newly collected data are fed back into the annotation and curation pipeline, forming a closed-loop process that continuously improves data coverage, diversity, and the quality of supervision.

\subsection{Dual Alignment for Heterogeneous Data}
\label{sec:dual_alignment}
The preceding preprocessing pipeline yields a clean and diverse corpus, but it does not resolve the mismatch in action supervision across data sources. Human videos often lack reliable physical action labels, while available human and robot trajectories are expressed in incompatible embodiment-specific representations. We therefore design a dual-alignment paradigm that routes data according to its available supervision: {implicit alignment} uses visual transitions to unify data without reliable action labels, whereas {explicit alignment} maps reliable human and robot trajectories into a shared physical action space.

\subsubsection{Implicit Alignment via Latent Actions}
\label{sec:implicit}
% Native actions are not directly comparable across embodiments, as the same physical effect may be produced by a human hand, a parallel gripper, or a dexterous robot hand using incompatible motor coordinates. However, such actions often induce similar object-centric visual transitions. We therefore infer shared latent actions from consecutive observations to provide transition-level conditioning for LAC-WM pretraining across human, simulation, and robot videos. This implicit alignment requires neither shared physical action labels nor paired cross-embodiment trajectories, relying instead on common motion, contact, and geometric structure.

Native actions are not directly comparable across embodiments, as the same physical effect may be produced by a human hand, a parallel gripper, or a dexterous robot hand using incompatible motor coordinates. Nevertheless, these actions often induce similar object-centric visual transitions. We therefore perform implicit alignment by inferring a shared latent-action space from consecutive observations across human, simulation, and robot videos. Relying on common motion, contact, and geometric structure, this process maps action-unlabeled or action-incompatible transitions into shared transition-level factors without requiring common physical action labels or paired cross-embodiment trajectories. We next describe the latent-action inference process, while their use as conditioning signals for LAC-WM pretraining is detailed in Sec.~\ref{sec:implicit-training}.

% Latent actions help structure LAC-WM learning. Predicting futures from only observations and instructions typically requires fitting a high-entropy mixture of transitions. Inferred latents resolve this mixture into lower-entropy conditional modes, mitigating gradient conflict. Learning these modes with shared parameters guides the backbone to model how scene structure responds to motion, contact, and geometric changes, supporting the learning of an action-sensitive dynamics manifold. Random condition dropout then connects this objective to the observation-language pathway: conditioned samples provide transition supervision, while dropped samples adapt the same dynamics to downstream inputs. The frozen LAC-WM consequently tends to expose more consistent dynamics features to the policy.
% I move this paragraph to the stage 1

\paragraph{Standardized multi-view observations.}
Let $\mathbf{I}^{v}_t$ denote the RGB image from view $v$ at time $t$, where the views follow the fixed order
$\mathcal{V}=(\mathrm{head},\mathrm{left},\mathrm{right})$.
We denote the uniformly preprocessed image by
$\widetilde{\mathbf{I}}^{v}_t$ and replace an unavailable wrist view with a zero image of the same size.
The three views are then spatially concatenated into a composite observation:
\begin{equation}
    \overline{\mathbf{o}}_t
    =
    \operatorname{SpatialConcat}\!\left(
        \widetilde{\mathbf{I}}^{\mathrm{head}}_t,
        \widetilde{\mathbf{I}}^{\mathrm{left}}_t,
        \widetilde{\mathbf{I}}^{\mathrm{right}}_t
    \right).
    \label{eq:implicit_multiview_observation}
\end{equation}
The latent-action model treats the resulting composite as a standard single-image input, providing a consistent observation format without source-specific view encoders.

\paragraph{Latent-action discovery.}
We train a shared variational latent-action model (LAM) across the multi-source video corpus, using consecutive observation pairs sampled independently from human, simulation, and robot trajectories. Its inverse dynamics encoder $q_{\eta}$ infers an abstract transition latent $\mathbf{z}_t$ from consecutive composite observations, while its forward transition decoder $T_{\xi}$ reconstructs the subsequent observation:
\begin{equation}
\begin{aligned}
    \mathbf{z}_t
    &\sim q_{\eta}
    (\mathbf{z}\mid\overline{\mathbf{o}}_t,
    \overline{\mathbf{o}}_{t+1}),\\
    \widehat{\overline{\mathbf{o}}}_{t+1}
    &=T_{\xi}(\overline{\mathbf{o}}_t,\mathbf{z}_t),
    \qquad \mathbf{z}_t\in\mathcal{Z}.
    \label{eq:implicit_latent_discovery}
\end{aligned}
\end{equation}
This asymmetric reconstruction design gives the latent a transition-specific role. Because the decoder is already conditioned on
$\overline{\mathbf{o}}_t$, it can access the static scene
content directly, encouraging $\mathbf{z}_t$ to encode the
transition-specific changes needed to reconstruct
$\overline{\mathbf{o}}_{t+1}$ rather than redundant appearance
information. The variational bottleneck makes it costly to spend latent capacity on incidental appearance or source identity, favoring a compact description of the residual transformation.
The LAM is trained with a combination of pixel-level, perceptual,
semantic, motion, and geometric reconstruction objectives, together
with a variational information bottleneck:
\begin{equation}
\mathcal{L}_{\mathrm{LAM}}
=
\mathbb{E}_{(
\overline{\mathbf{o}}_t,
\overline{\mathbf{o}}_{t+1})
\sim\pi_{\mathcal{D}}}
\left[
\sum_{r\in\mathcal{R}}
\lambda_r
\ell_r\!\left(
\widehat{\overline{\mathbf{o}}}_{t+1},
\overline{\mathbf{o}}_{t+1}
\right)
+
\beta_{\mathrm{KL}}
D_{\mathrm{KL}}\!\left(
q_{\eta}\!\left(
\mathbf{z}\mid
\overline{\mathbf{o}}_t,
\overline{\mathbf{o}}_{t+1}
\right)
\,\middle\|\,
p(\mathbf{z})
\right)
\right].
\label{eq:implicit_lam_objective}
\end{equation}

Here, $\mathcal{R}
=
\{\mathrm{L1},\mathrm{LPIPS},\mathrm{DINO},
\mathrm{flow},\mathrm{depth},\mathrm{VGGT}\}$
denotes the set of reconstruction criteria, $\pi_{\mathcal{D}}$ denotes the sampling distribution over
transitions from the multi-source corpus, and
$\ell_r$ denotes the reconstruction criterion associated with
$r\in\mathcal{R}$. The coefficient $\lambda_r$ controls the
contribution of each reconstruction term, while
$\beta_{\mathrm{KL}}$ weights the divergence between the inferred
posterior and the latent prior $p(\mathbf{z})$. The reconstruction
terms compare the predicted and ground-truth next observations either
directly or through their derived features. Specifically, L1 and LPIPS~\citep{zhang2018lpips}
preserve pixel-level and perceptual appearance, DINO~\citep{caron2021dino} maintains
high-level semantic consistency, optical flow captures motion, and
depth and VGGT~\citep{wang2025vggt} provide geometric supervision.

Sharing the encoder, decoder, and latent prior across data sources, together with the motion and geometric objectives, encourages recurring physical transitions to reuse common latent factors rather than encode source-specific appearance. In this setting, learning a shared latent-action space constitutes implicit alignment: source-specific visual transitions are mapped into common transition-level factors without paired trajectories or shared physical action labels. After latent-action discovery, we freeze $q_{\eta}$ and use its posterior mean $\bar{\mathbf{z}}_t=\mathbb{E}_{q_{\eta}}[\mathbf{z}\mid\overline{\mathbf{o}}_t,\overline{\mathbf{o}}_{t+1}]$ as the deterministic offline label for each human, simulation, and robot transition. These inferred latent actions provide scalable transition-level conditioning for the LAC-WM pretraining described in Sec.~\ref{sec:implicit-training}. The decoder $T_{\xi}$ is discarded after latent-action discovery and is not used in subsequent stages.

% 之浩
\subsubsection{Explicit Alignment via Unified Physical Action Space}
\label{sec:explicit}
% Implicit alignment enables action-unlabeled videos to provide transition-level supervision, but the resulting latent actions are not grounded in an executable physical control space. For human and robot trajectories with reliable motion or action annotations, we therefore introduce explicit alignment, which maps their heterogeneous native representations into a shared canonical space with fixed physical semantics. This alignment allows reliable physical trajectories from different embodiments to provide consistent supervision while retaining only the dimensions valid for each embodiment.

Implicit alignment lets action-unlabeled videos provide transition-level dynamics supervision, whereas executable control additionally requires supervision defined directly in a unified physical action space. For human and robot trajectories that carry reliable motion or action annotations, we therefore design explicit alignment to turn their heterogeneous native formats into physically consistent, cross-embodiment supervision through a 130-D canonical state-action representation and camera-frame chunk-relative end-effector actions.

\begin{table}[htbp]
\centering
\caption{
\textbf{Detailed layout of the 130-dimensional canonical state-action representation.}
Rows and index ranges are ordered by side and body; ranges use the half-open convention $[s,e)$.
}
\label{tab:canonical_representation}
\small
\setlength{\tabcolsep}{4pt}
\renewcommand{\arraystretch}{1.08}
\begin{tabularx}{\linewidth}{@{} l l c c X @{}}
\toprule
\textbf{Block}
& \textbf{Canonical slot}
& \textbf{Range}
& \textbf{Dim.}
& \textbf{Physical semantics} \\
\midrule

Left
& Arm joints
& $[0,7)$
& 7
& Up to seven left-arm joint positions. \\

& End-effector pose
& $[7,13)$
& 6
& Cartesian position (3) and axis-angle rotation (3). \\

& Gripper
& $[13,14)$
& 1
& Parallel-gripper position or opening width. \\

& Fingertip positions
& $[14,29)$
& 15
& Five fingertip XYZ positions in the order thumb, index, middle, ring, and little finger. \\

& Hand joints
& $[29,54)$
& 25
& Up to 25 dexterous-hand joint positions. \\

\midrule

Right
& Arm joints
& $[54,61)$
& 7
& Up to seven right-arm joint positions. \\

& End-effector pose
& $[61,67)$
& 6
& Cartesian position (3) and axis-angle rotation (3). \\

& Gripper
& $[67,68)$
& 1
& Parallel-gripper position or opening width. \\

& Fingertip positions
& $[68,83)$
& 15
& Five fingertip XYZ positions in the order thumb, index, middle, ring, and little finger. \\

& Hand joints
& $[83,108)$
& 25
& Up to 25 dexterous-hand joint positions. \\

\midrule

Body
& Base velocity
& $[108,114)$
& 6
& Linear and angular base velocity. \\

& Head pose
& $[114,118)$
& 4
& Head-related pose or control variables. \\

& Waist pose
& $[118,122)$
& 4
& Waist-related pose or control variables. \\

& Reserved
& $[122,130)$
& 8
& Reserved for additional embodiment-specific variables. \\

\bottomrule
\end{tabularx}
\end{table}

\paragraph{130-D canonical state-action representation.}
To provide physically grounded supervision across human egocentric, simulation, and real-robot data, we map trajectories with reliable motion or action annotations into a shared 130-dimensional representation. As detailed in Table~\ref{tab:canonical_representation}, the representation is organized into left, right, and body blocks, with each canonical slot assigned a fixed physical meaning. The left and right blocks represent arm joints, end-effector poses, grippers, Cartesian fingertips, and dexterous-hand joints, while the body block represents shared body states. For each source or embodiment $e$, an adaptor $\Phi_e$ maps its native state or action sequence into the canonical space:
\begin{equation}
\widetilde{\mathbf{X}}_t
=
\Phi_e\!\left(\mathbf{X}^{(e)}_t\right)
\in
\mathbb{R}^{H_X \times 130},
\qquad
\mathbf{X}\in\{\mathbf{S},\mathbf{A}\},
\label{eq:canonical_mapping}
\end{equation}
where $H_X$ denotes the corresponding state or action horizon. For brevity, we omit the tilde from canonicalized states and actions in subsequent sections. Each adapter standardizes units, coordinate-frame conventions, rotation representations, and semantic ordering before assigning available measurements to their canonical slots. For simulation and real-robot trajectories, joint, end-effector, gripper, base, and dexterous-hand signals are written into the corresponding slots. For human trajectories, we avoid embodiment-specific retargeting and dense MANO parameterization by using a sparse task-space representation in which each wrist pose occupies an end-effector slot and the five ordered fingertip XYZ coordinates occupy the corresponding 15-dimensional fingertip-position slot. End-effector orientations are represented in axis-angle form, while unsupported dimensions are zero-filled and excluded from training by a validity mask.

\paragraph{Camera-frame chunk-relative end-effector action.}
To enable the transfer of trajectory-level knowledge across egocentric, simulation, and real-robot data, we convert wrist or end-effector channels with pose semantics into a camera-aligned, chunk-relative action representation. The camera frame provides a common observation-centric convention across embodiments and directly aligns the action representation with the policy's visual input, reducing reliance on robot-specific base or world coordinates. Expressing the targets relative to the conditioning end-effector state further reduces sensitivity to absolute pose and calibration differences while preserving the underlying manipulation motion, thereby weakening embodiment dependence and promoting generalist policy learning.

% For an action chunk conditioned on the state at time $t$, let $\{\kappa_i\}_{i=0}^{H-1}$ denote the timestamps of its $H$ target poses. Let $E_k$ denote the coordinate frame attached to the human wrist or robot end effector at time $k$, and let ${}^{A}\mathbf{T}_{B}$ denote the rigid transformation from frame $B$ to frame $A$. Through source-specific coordinate transformations, the conditioning pose and all target poses are first expressed in the camera frame $C_t$ associated with the conditioning observation. We denote the conditioning pose by ${}^{C_t}\mathbf{T}^{\mathrm{state}}_{E_t}$ and the aligned target poses by ${}^{C_t}\widehat{\mathbf{T}}^{\mathrm{tgt}}_{E_{\kappa_i}}$. Each target is then expressed relative to the same conditioning end-effector frame:
% \begin{equation}
% {}^{E_t}\mathbf{T}^{\mathrm{rel}}_{E_{\kappa_i}}
% =
% \left({}^{C_t}\mathbf{T}^{\mathrm{state}}_{E_t}\right)^{-1}
% {}^{C_t}\widehat{\mathbf{T}}^{\mathrm{tgt}}_{E_{\kappa_i}},
% \qquad i=0,\ldots,H-1.
% \label{eq:camera_frame_relative_action}
% \end{equation}
% The resulting transformation describes the pose of each target end-effector frame $E_{\kappa_i}$ relative to the conditioning frame $E_t$. Thus, all targets in the chunk share the same reference rather than being represented as frame-to-frame differences. The translation and axis-angle rotation of ${}^{E_t}\mathbf{T}^{\mathrm{rel}}_{E_{\kappa_i}}$ form the six-dimensional end-effector action, and the axis-angle trajectory is temporally unwrapped before normalization.
Consider an action chunk conditioned on the state at time $t$, comprising $H$ end-effector pose targets at timestamps $\{t_i\}_{i=0}^{H-1}$. We denote the conditioning and target end-effector frames by $E_t$ and $E_{t_i}$, and their corresponding camera frames by $C_t$ and $C_{t_i}$, respectively. We use ${}^{A}\mathbf{T}_{B}$ to denote the rigid transformation from frame $B$ to frame $A$. After source-specific coordinate conversion, the conditioning pose is represented by ${}^{C_t}\mathbf{T}^{\mathrm{state}}_{E_t}$, while the $i$-th target pose is represented by ${}^{C_{t_i}}\mathbf{T}^{\mathrm{tgt}}_{E_{t_i}}$ in its instantaneous camera frame. Using a fixed trajectory-level reference frame $W$ to relate camera poses across timestamps, we first transform each target pose into the conditioning camera frame $C_t$ and then express it relative to the conditioning end-effector state:
\begin{equation}
\begin{aligned}
{}^{C_t}\widehat{\mathbf{T}}^{\mathrm{tgt}}_{E_{t_i}}
&=
\left({}^{W}\mathbf{T}_{C_t}\right)^{-1}
{}^{W}\mathbf{T}_{C_{t_i}}
{}^{C_{t_i}}\mathbf{T}^{\mathrm{tgt}}_{E_{t_i}}, \\
{}^{E_t}\mathbf{T}^{\mathrm{rel}}_{E_{t_i}}
&=
\left({}^{C_t}\mathbf{T}^{\mathrm{state}}_{E_t}\right)^{-1}
{}^{C_t}\widehat{\mathbf{T}}^{\mathrm{tgt}}_{E_{t_i}},
\qquad i=0,\ldots,H-1.
\end{aligned}
\label{eq:camera_frame_relative_action}
\end{equation}
The first transformation maps each target pose from $C_{t_i}$ into the fixed conditioning camera frame $C_t$, thereby compensating for camera motion within the chunk. When the camera remains fixed, this transformation reduces to the identity. The second transformation then expresses every aligned target relative to the same conditioning end-effector frame $E_t$. Thus, the targets share a common reference across the entire chunk rather than being represented as frame-to-frame differences.

Together, canonical slot mapping and camera-aligned chunk-relative actions provide physically consistent supervision across heterogeneous trajectories. At deployment, only the canonical slots required by the target robot are extracted and converted into its native control commands.

\section{Approach}
\label{sec:pipeline}
\begin{figure}[!tbp]
    
    \includegraphics[width=\linewidth]{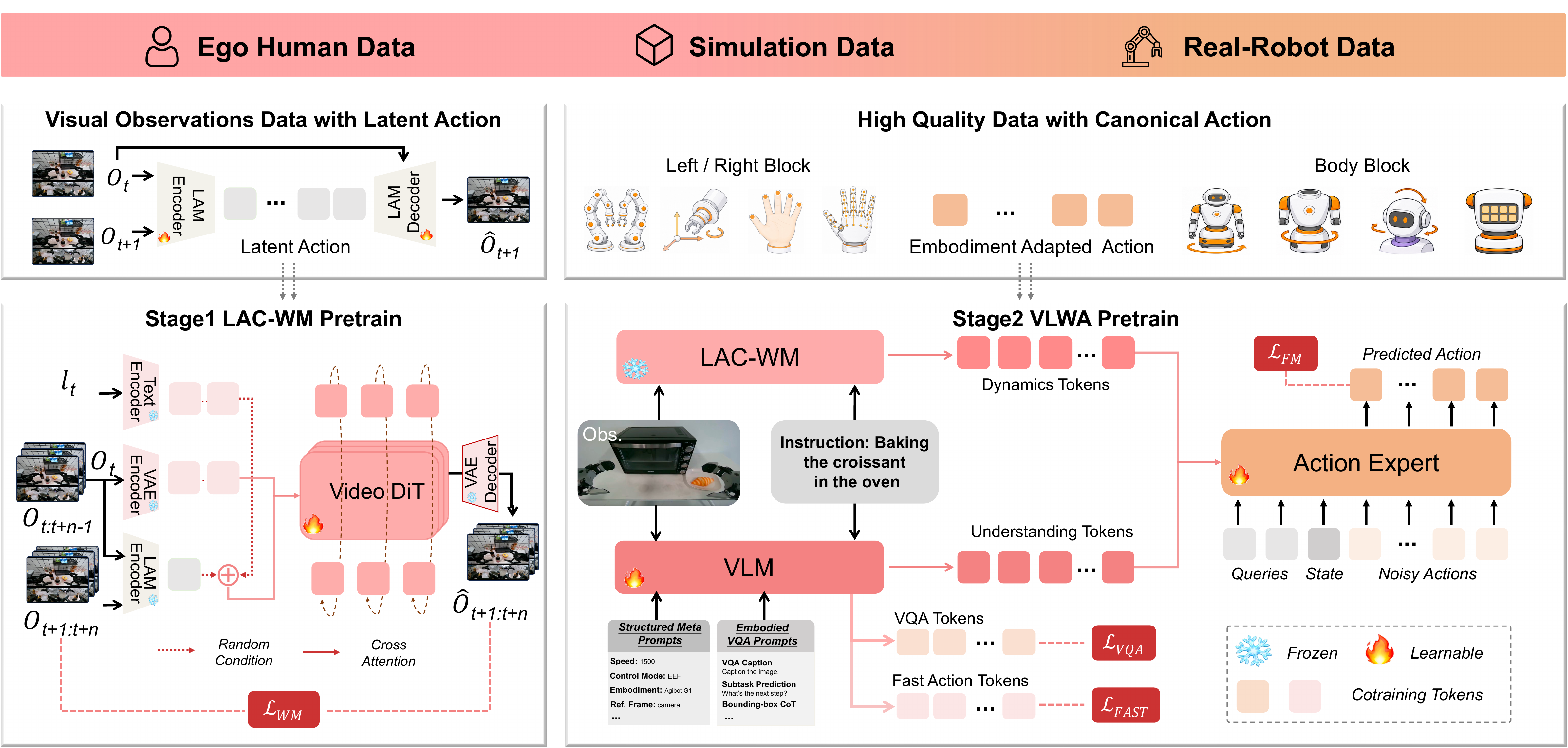}
    \centering
    \caption{\textbf{Overview of the two-stage pretraining framework.} Implicit alignment learns transferable dynamics from heterogeneous videos through latent-action-conditioned world-model pretraining, while explicit alignment maps reliable human and robot trajectories into a unified action space. The VLM and LAC-WM provide complementary semantic and dynamics representations that jointly condition the action expert for executable robot control.
    } \label{fig:model_architecture}
\end{figure}

As illustrated in Figure~\ref{fig:model_architecture}, JoyAI-RA 0.5 is a Vision-Language-World-Action framework designed to learn generalist manipulation policies from heterogeneous data. Sec.~\ref{sec:model_architecture} introduces its three-component architecture, where a VLM and a Latent-Action-Conditioned WM provide complementary semantic and dynamics representations for a Flow-Matching Action Expert that generates continuous actions. Sec.~\ref{sec:training} then describes the staged training process, consisting of LAC-WM pretraining, VLWA pretraining, and target-robot post-training. Finally, Sec.~\ref{sec:rl} presents an inner-outer-loop reinforcement-learning framework for efficient task-specific adaptation and continual improvement of the foundation policy.

% 达丰
\subsection{Model Architecture}
\label{sec:model_architecture}

At control step $t$, the model takes as input a multi-view visual observation
$\mathbf{o}_t=\{\mathbf{I}^{v}_t\}_{v=1}^{V}$, a language instruction $\ell$, and the current proprioceptive state $\mathbf{s}_t$. It predicts an $H$-step action chunk
\begin{equation}
    \mathbf{A}_t
    =
    [\mathbf{a}_t,\ldots,\mathbf{a}_{t+H-1}]
    \in
    \mathbb{R}^{H\times D_a},
    \label{eq:action_chunk}
\end{equation}
where $D_a=130$ is the dimension of the canonical cross-embodiment action space introduced in Sec.~\ref{sec:explicit}.

The visual observation and instruction are processed independently by the VLM and LAC-WM. The VLM produces a sequence of semantic representations
$\mathbf{U}_t\in\mathbb{R}^{L_V\times d}$ that encodes the task objective, relevant objects, spatial relationships, and their correspondence with the instruction. In parallel, the LAC-WM produces dynamics representations
$\mathbf{D}_t\in\mathbb{R}^{L_W\times d}$ that capture regularities in how manipulation scenes evolve. It acquires this transition knowledge through latent-action-conditioned pretraining, using the latent actions inferred in Sec.~\ref{sec:implicit} under the training procedure described in Sec.~\ref{sec:implicit-training}.

The semantic and dynamics representations are concatenated along the sequence dimension:
\begin{equation}
    \mathbf{C}_t
    =
    [\mathbf{U}_t;\mathbf{D}_t]
    \in
    \mathbb{R}^{(L_V+L_W)\times d}.
    \label{eq:architecture_context}
\end{equation}
This late-fusion design preserves the specialization of the two backbones while exposing both sources of information to the action expert. The VLM primarily specifies {what} outcome should be achieved, whereas the LAC-WM provides a learned prior about {how} the observed scene may evolve during interaction.

The action expert is an attention-based flow-matching network that converts the fused context into continuous control~\cite{zhang2026joyai}. At flow time $\tau\in[0,1]$, it receives a noisy action 
$\mathbf{X}_{t,\tau}\in\mathbb{R}^{H\times D_a}$ together with the proprioceptive state, flow-time embedding, and learnable action queries. The action-side features attend to $\mathbf{C}_t$ and predict the conditional velocity field
\begin{equation}
    \widehat{\mathbf{v}}_{t,\tau}
    =
    F_{\theta_A}
    \left(
        \mathbf{X}_{t,\tau},
        \tau,
        \mathbf{s}_t,
        \mathbf{C}_t
    \right)
    \in
    \mathbb{R}^{H\times D_a}.
    \label{eq:action_expert}
\end{equation}
At inference, the action state is initialized from Gaussian noise and progressively integrated according to the predicted velocity field to obtain the final action chunk $\widehat{\mathbf{A}}_t$.

% 达丰
\subsection{Training Paradigm}
\label{sec:training}
JoyAI-RA 0.5 adopts a four-stage training paradigm that progressively converts heterogeneous video and action data into a deployment-ready robot policy. Stage~1 pretrains the LAC-WM on implicitly aligned videos to acquire general interaction dynamics (Sec.~\ref{sec:implicit-training}). Stage~2 performs cross-embodiment VLWA pretraining on explicitly aligned trajectories, jointly optimizing the VLM and Action Expert while keeping the LAC-WM frozen (Sec.~\ref{sec:vlwa-pretraining}). Stage~3 adapts the policy to the target robot using high-quality, deployment-relevant demonstrations (Sec.~\ref{sec:post-training}). Finally, Stage~4 (Sec.~\ref{sec:rl}) employs inner-outer loop reinforcement learning to further improve task-specific performance and generalization. Stages~1--2 constitute pretraining, whereas Stages~3--4 comprise post-training.

\subsubsection{Stage 1: LAC-WM Pretraining}\label{sec:implicit-training}
% \paragraph{Pretraining the LAC-WM.}

Building on the implicit alignment introduced in Sec.~\ref{sec:implicit}, we use the inferred latent actions as transition-level conditioning signals to pretrain the LAC-WM, denoted by $W_{\theta_W}$. For a prediction horizon of $H_v$, we collect the corresponding offline labels into the latent-action chunk $\bar{\mathbf{Z}}_t=[\bar{\mathbf{z}}_t,\ldots,\bar{\mathbf{z}}_{t+H_v-1}]$. Depending on the available annotations, we sample $\mathbf{c}_t\sim\pi_c$ from the available elements of $\{\ell,\bar{\mathbf{Z}}_t,(\ell,\bar{\mathbf{Z}}_t)\}$ and predict
\begin{equation}
    \widehat{\mathbf{O}}^{\mathrm{future}}_t=W_{\theta_W}(\mathbf{o}_t;\mathbf{c}_t),\qquad \mathbf{O}^{\mathrm{future}}_t=\mathbf{o}_{t+1:t+H_v}.
    \label{eq:implicit_lacwm_prediction}
\end{equation}
% Here, $H_v$ is the prediction horizon and $\pi_c$ samples from the available conditioning modes. For videos without language annotations, the inferred latent action $\mathbf{z}_t$ provides the available conditioning signal during LAC-WM pretraining. Latent actions provide local transition cues, language specifies the task, and joint conditioning connects visual dynamics with instructions when both are available. During conditioning-mode sampling, we replace $\mathbf{z}_t$ with a null condition with probability $p_{\mathrm{drop}}$, jointly training transition-conditioned prediction and the downstream observation-language pathway. All sources and conditioning modes update the same parameters $\theta_W$.
Here, $H_v$ denotes the prediction horizon, and $\pi_c$ selects among the conditioning modes available for each training sample. The latent-action chunk $\bar{\mathbf{Z}}_t$ captures the sequence of local transitions over that horizon, the language instruction $\ell$ provides task semantics, and their joint use associates these semantics with visual dynamics. For videos without language annotations, $\bar{\mathbf{Z}}_t$ serves as the available conditioning signal. To support subsequent stages without latent-action inference, whenever the sampled mode contains $\bar{\mathbf{Z}}_t$, we replace it with a null condition with probability $p_{\mathrm{drop}}$. This condition-dropout strategy jointly trains latent-action-conditioned and latent-action-free prediction across all data sources using the same parameters $\theta_W$.

This conditioning design also structures LAC-WM representation learning. Without transition-level conditioning, prediction from observations and instructions must account for a high-entropy mixture of plausible futures. Latent actions decompose this mixture into more coherent transition-specific modes, reducing prediction ambiguity and gradient interference. Learning these modes with a shared backbone encourages the model to capture how scenes respond to motion, contact, and geometric changes, while condition dropout transfers the resulting action-sensitive dynamics structure to the observation-language pathway used downstream.

Following DreamDojo~\citep{gao2026dreamdojo}, we optimize the LAC-WM with flow-matching velocity regression and a temporal-difference term. Let $\mathbf{y}$ denote the ground-truth future-video latent, $\boldsymbol{\epsilon}\sim\mathcal{N}(\mathbf{0},\mathbf{I})$, and $\tau\sim\mathcal{U}(0,1)$. We construct
\begin{equation}
    \mathbf{x}_{\tau}
    =
    (1-\tau)\boldsymbol{\epsilon}
    +
    \tau\mathbf{y},
    \qquad
    \mathbf{v}^{\star}_{\tau}
    =
    \mathbf{y}-\boldsymbol{\epsilon}.
    \label{eq:flow_matching_path}
\end{equation}
Let $K$ denote the number of temporal positions in the future-video latent, and let $\mathbf{v}^{\star}_{\tau,i}$ denote the target velocity at position $i$. The predicted velocity is $\widehat{\mathbf{v}}_{\tau,i}=u_{\theta_W}(\mathbf{x}_{\tau},\tau,\mathbf{o}_t,\mathbf{c}_t)_i$:
\begin{equation}
    \mathcal{L}^{\mathrm{pre}}_{\mathrm{video}}=\mathbb{E}\!\left[\sum_{i=1}^{K}\|\widehat{\mathbf{v}}_{\tau,i}-\mathbf{v}^{\star}_{\tau,i}\|_2^2+\lambda_{\mathrm{temporal}}\sum_{i=1}^{K-1}\|\Delta\widehat{\mathbf{v}}_{\tau,i}-\Delta\mathbf{v}^{\star}_{\tau,i}\|_2^2\right].
    \label{eq:implicit_lacwm_objective}
\end{equation}
where $\Delta\widehat{\mathbf{v}}_{\tau,i}=\widehat{\mathbf{v}}_{\tau,i+1}-\widehat{\mathbf{v}}_{\tau,i}$ and $\Delta\mathbf{v}^{\star}_{\tau,i}=\mathbf{v}^{\star}_{\tau,i+1}-\mathbf{v}^{\star}_{\tau,i}$. The first term matches the target velocity at each latent position, whereas the second matches velocity differences between adjacent positions, improving temporal consistency in future-video prediction and action-following fidelity.

During subsequent VLWA pretraining and deployment, the LAC-WM is frozen and run causally with the current observation as the first frame together with the language instruction. We extract the hidden features associated with this first frame:
\begin{equation}
    \mathbf{D}_t
    =W^{\mathrm{feat}}_{\theta_W}
    (\mathbf{o}_t,\ell).
    \label{eq:implicit_downstream_features}
\end{equation}
Here, $W^{\mathrm{feat}}_{\theta_W}$ denotes causal first-frame feature extraction from the LAC-WM.
Because the conditioned and dropped objectives share $\theta_W$, $\mathbf{D}_t$ retains the action-sensitive dynamics structure learned by disambiguating future transition modes.
The Action Expert consumes $\mathbf{D}_t$ as conditioning context, transferring dynamics learned from action-unlabeled videos to executable action prediction.

\subsubsection{Stage 2: VLWA Pretraining}\label{sec:vlwa-pretraining}
Stage~2 uses action-annotated robot and simulation trajectories, together with egocentric trajectories whose hand motions can be reliably recovered.
As described in Sec.~\ref{sec:explicit}, each native action chunk
$\mathbf{A}^{(e)}_t$
is mapped by an embodiment-specific adaptor
$\Phi_e$
into the canonical action space:
\begin{equation}
    \mathbf{A}_t
    =
    \Phi_e
    \left(
        \mathbf{A}^{(e)}_t
    \right)
    \in
    \mathbb{R}^{H\times D_a},
    \qquad
    \mathbf{M}_t
    \in
    \{0,1\}^{H\times D_a},
    \label{eq:explicit_action_alignment}
\end{equation}
where $\mathbf{M}_t$ marks the action dimensions supported by embodiment $e$.
Egocentric clips without reliable physical trajectories remain available for Stage~1 but do not receive explicit action supervision in this stage.

For each training sample, the frozen LAC-WM extracts a dynamics representation from the current observation and language instruction, while the VLM produces the corresponding task-directed semantic representation.
The action expert attends to their fused context and learns continuous control through masked flow matching~\cite{lipman2022flow}.
We reuse the linear flow path defined in Eq.~(\ref{eq:flow_matching_path}), replacing the future-video latent $\mathbf{y}$ with the canonical action target $\mathbf{A}_t$ and sampling $\boldsymbol{\epsilon}_t\sim\mathcal{N}(\mathbf{0},\mathbf{I})$. This yields the noisy action $\mathbf{X}_{t,\tau}$ and target velocity $\mathbf{v}^{\star}_{t,\tau}$, and we optimize
\begin{equation}
    \mathcal{L}_{\mathrm{FM}}
    =
    \mathbb{E}_{\tau,\boldsymbol{\epsilon}_t}
    \left[
        \frac{
            \left\|
                \mathbf{M}_t
                \odot
                \left(
                    \widehat{\mathbf{v}}_{t,\tau}
                    -
                    \mathbf{v}^{\star}_{t,\tau}
                \right)
            \right\|_2^2
        }{
            \|\mathbf{M}_t\|_1
        }
    \right].
    \label{eq:masked_flow_matching}
\end{equation}
The mask prevents unsupported action dimensions from contributing to the objective, allowing heterogeneous embodiments to share the same action expert.

In addition to continuous action learning, the VLM is optimized with a VQA objective that strengthens semantic understanding, spatial grounding, and subtask reasoning, together with a FAST objective derived from the unified action representation that provides action-aware supervision~\cite{pertsch2025pi0fast}.
The Stage~2 objective is
\begin{equation}
    \mathcal{L}_{\mathrm{stage2}}
    =
    \lambda_{\mathrm{VQA}}\mathcal{L}_{\mathrm{VQA}}
    +
    \lambda_{\mathrm{FAST}}\mathcal{L}_{\mathrm{FAST}}
    +
    \lambda_{\mathrm{FM}}\mathcal{L}_{\mathrm{FM}},
    \label{eq:stage2_objective}
\end{equation}
where each term is evaluated only on samples with the corresponding supervision.
The VLM and action expert are optimized jointly, whereas the LAC-WM remains frozen. By keeping the LAC-WM frozen, Stage~2 preserves its pretrained dynamics representations while jointly adapting the VLM and action expert to executable control.
This decouples dynamics acquisition from action grounding and provides a stable foundation for cross-embodiment policy learning.

\subsubsection{Stage 3: Target-Robot Post-Training}\label{sec:post-training}
Starting from the cross-embodiment policy learned in Stage~2, we further post-train JoyAI-RA on high-quality demonstrations collected from the target robot and deployment-relevant tasks.
We follow the same training formulation as in Stage~2, using the flow-matching supervision, while restricting the training data and valid action dimensions to the target embodiment.
The LAC-WM remains frozen, whereas the VLM and action expert are adapted to the target robot's visual observations, kinematics, and control interface.
This stage transforms the broadly pretrained policy into a deployment-ready target-robot policy and provides a strong initialization for the subsequent reinforcement-learning stage.

\subsection{Inner-Outer Loop Reinforcement Learning}
\label{sec:rl}

\begin{figure}[!tbp]
\centering
\includegraphics[width=1.0\linewidth]{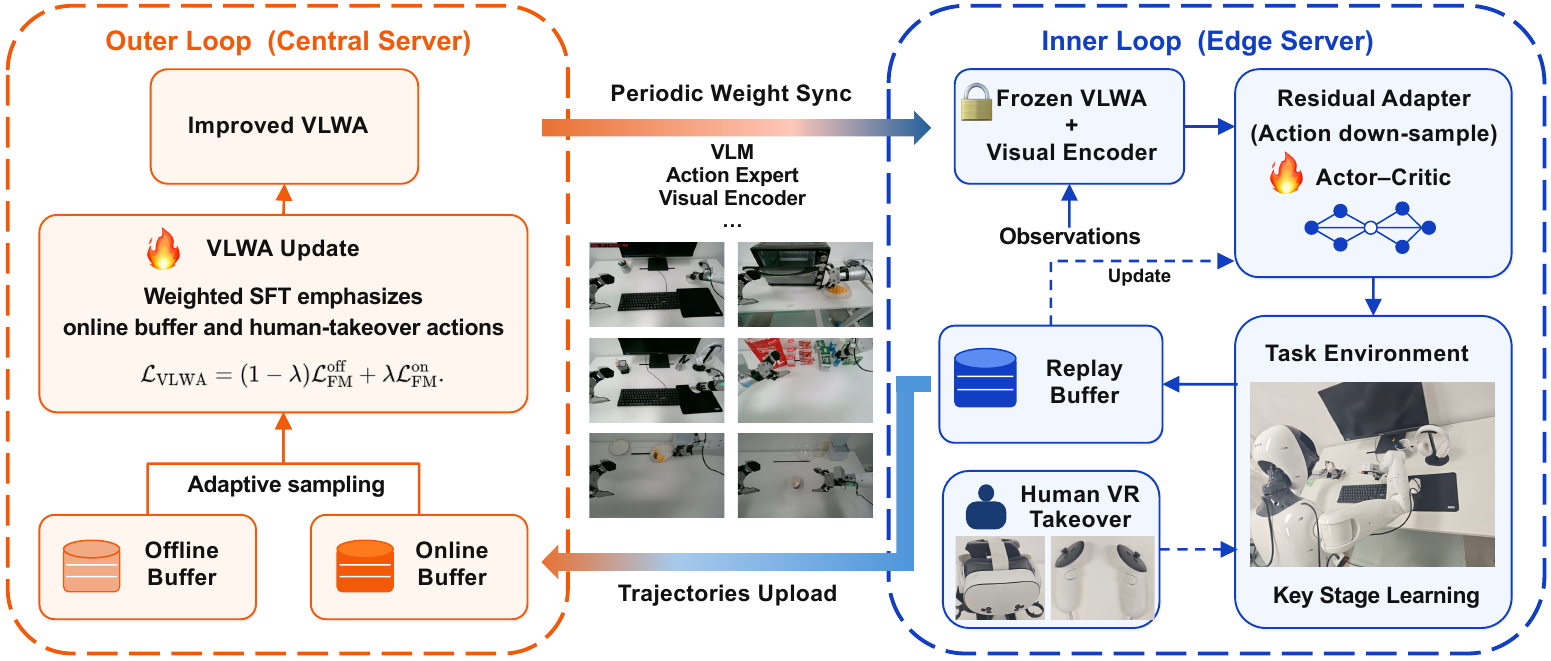}
\caption{
\textbf{Overview of the inner-outer loop reinforcement learning framework.}
The inner loop performs efficient task-specific adaptation on the edge server, while the asynchronous outer loop improves the VLWA model on the central server and periodically synchronizes the updated parameters back, forming a closed self-improving loop.
}
\label{fig:rl_inner_outer_loop}
\end{figure}

% To couple fast task-specific adaptation with continual foundation-policy improvement during deployment, we design an inner-outer loop reinforcement learning stage. As shown in Fig.~\ref{fig:rl_inner_outer_loop}, a fast inner loop on the edge server adapts to the current task, while an asynchronous outer loop on the central server refines the foundation VLWA and periodically synchronizes it back, so the two loops operate at different timescales yet reinforce each other.
To combine rapid task adaptation with continual foundation-policy improvement, we introduce an inner--outer loop reinforcement learning stage. As shown in Fig.~\ref{fig:rl_inner_outer_loop}, a fast inner loop on the edge server adapts to the current task, while an asynchronous outer loop on the central server refines the foundation VLWA and periodically synchronizes its updates to the edge. Operating at different timescales, the two loops reinforce each other.

Specifically, the inner loop targets efficient task-specific adaptation. Rather than updating the entire VLWA, we freeze the foundation model and train a lightweight adapter that maps the VLWA reference action and visual embedding into a residual policy, using an efficient off-policy RL algorithm~\cite{xu2026rltokenbootstrappingonline} with human-in-the-loop interaction. Optimizing this residual in a downsampled low-dimensional space and updating only at critical stages yields rapid gains on in-domain tasks at low cost. However, with the foundation model and visual encoder frozen, the residual policy is confined to the interaction distribution seen during post-training and generalizes poorly beyond it. The asynchronous outer loop lifts this ceiling by improving the foundation model itself. Successful trajectories from inner-loop interaction—both autonomous rollouts and human interventions—are aggregated with the original post-training data to update the VLWA, and the improved parameters are periodically synchronized back so that the residual policy adapts on top of an increasingly capable foundation. To keep off-policy optimization stable across such updates, we refresh the replay buffer and warm up the actor-critic networks before resuming residual training.

Overall, the two loops form a closed, self-improving system: the outer loop steadily strengthens the generalization of the foundation VLWA by accumulating interaction experience, while the inner loop enables rapid task-specific adaptation based on the evolving foundation model, jointly improving policy performance.

\section{Experiments}
\label{sec:experiments}

\subsection{Experimental Setup}
\label{sec:exp_setup}

% As shown in Figure~\ref{fig:exp-settings}, we evaluate JoyAI-RA on real-world benchmarks to assess its effectiveness and generalization.

\begin{figure}[h]
    \centering
    \includegraphics[width=0.95\linewidth]{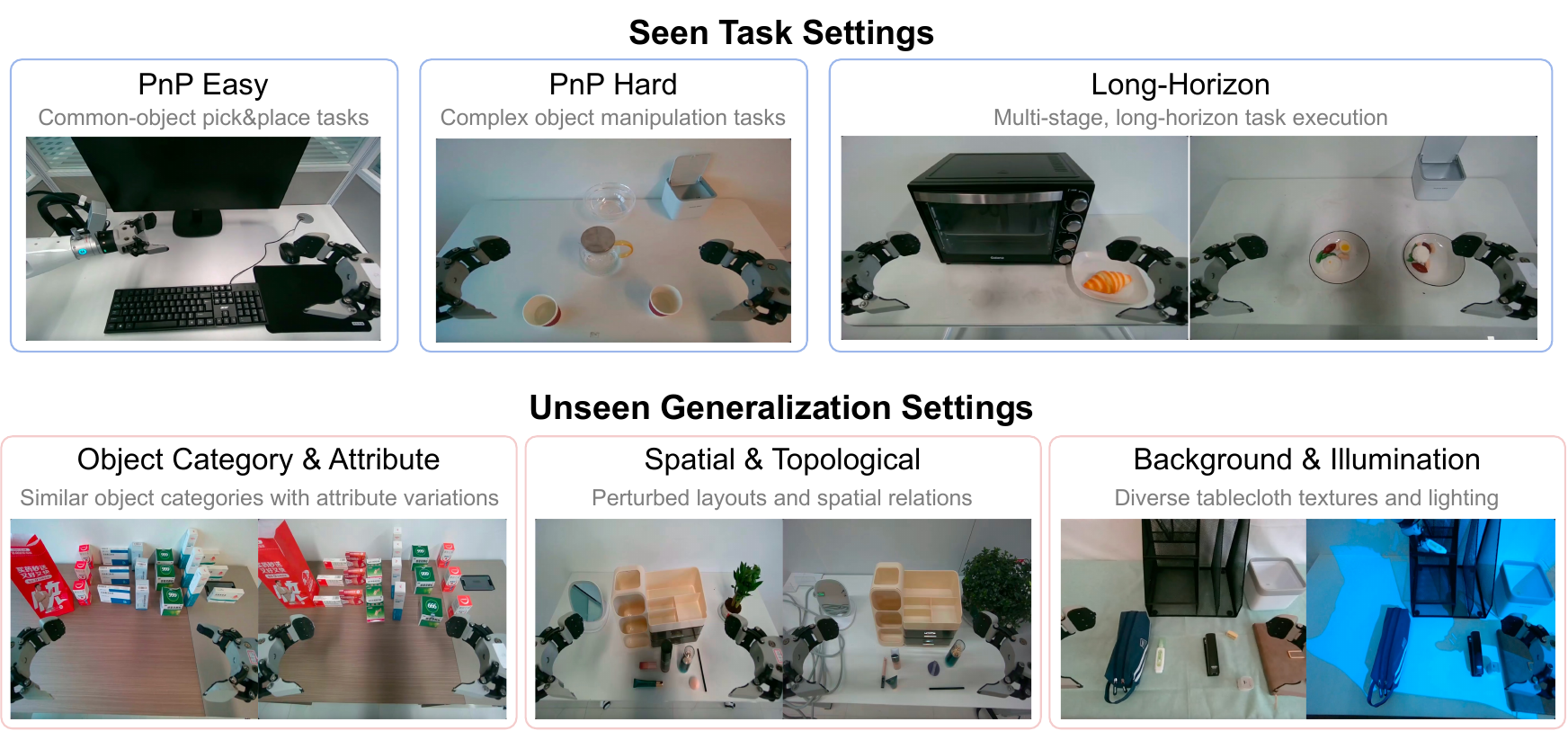}
    \caption{\textbf{Overview of the task and generalization settings.}}
    \label{fig:exp-settings}
\end{figure}

As shown in Figure~\ref{fig:exp-settings}, to systematically assess the real-world manipulation capabilities of JoyAI-RA, we establish the Real-World AgiBot Benchmark on the AgiBot G1 robotic platform. This benchmark encompasses three major task categories across six representative scenarios: office, tea room, kitchen, dining table, pharmacy, and dressing table environments. We organize the evaluation into two settings: seen performance, which measures task execution under familiar conditions, and unseen generalization, which evaluates robustness to novel objects, spatial and topological configurations, backgrounds, and illumination conditions.

% \begin{itemize}
% \item \textbf{PnP-Easy}: Pick-and-place tasks involving common objects (e.g., mouse and stapler) and organizing items (e.g., placing erasers or chargers into storage boxes);
% \item \textbf{PnP-Hard}: High-precision or complex manipulation tasks, such as picking and placing headphones, pens or correction fluid, and clearing food to organize plates;
% \item \textbf{Long-Horizon Tasks}: Multi-stage, sequential operations including Cup Discarding, Croissant Toasting, Remedy Packaging.
% \end{itemize}
Seen performance covers three task categories. \textbf{PnP-Easy} includes pick-and-place tasks involving common objects (e.g., mouse and stapler) and organizing items (e.g., placing erasers or chargers into storage boxes). \textbf{PnP-Hard} focuses on high-precision or complex manipulation tasks, such as picking and placing headphones, pens, or correction fluid, and cup discarding. \textbf{Long-Horizon Tasks} involve multi-stage, sequential operations, including clearing food to organize plates, croissant toasting, and remedy packaging.

% Building upon this foundation, we further evaluate the generalization capabilities of our model in unseen settings, focusing primarily on the following three dimensions:
% \begin{itemize}
% \item \textbf{Object Category and Attribute Generalization}: Evaluating performance using structurally similar categories or variations within the same category across different colors, dimensions, and visual appearances.
% \item \textbf{Spatial and Topological Generalization}: Introducing large-scale perturbations to the initial positions of objects and altering their spatial topological configurations relative to other entities.
% \item \textbf{Background and Illumination Generalization}: Modifying the tabletop environment via diverse tablecloth textures alongside substantial variations in ambient lighting conditions.
% \end{itemize}
For unseen generalization, we evaluate three dimensions. \textbf{Spatial and Topological Generalization (STG)} introduces large-scale perturbations to the initial positions of objects and alters their spatial topological configurations relative to other entities. \textbf{Object Category and Attribute Generalization (OCAG)} evaluates performance using structurally similar categories or variations within the same category across different colors, dimensions, and visual appearances. \textbf{Background and Illumination Generalization (BIG)} introduces diverse tablecloth textures and substantial variations in ambient lighting conditions.

For each task, we conduct 20 seen and 10 unseen trials. We report the mean task score on a 100-point scale, calculated from subtask completion rates across trials.
% , with the latter split into 2, 4, and 4 trials for OCAG, STG, and BIG, respectively.
% ====================================================================
% 请确保您的导言区(\begin{document} 之前)引入了以下宏包：
% \usepackage{booktabs}
% \usepackage{multicol}
% ====================================================================

% \begin{table}[H]
% \centering
% \footnotesize % 使用更小、更精致的学术字号，确保双栏排版不超界
% \setlength{\tabcolsep}{3.5pt} % 优化列间距，防止文字溢出
% \caption{Quantitative comparison of success rates (\%) between seen performance and unseen generalization. Best results are highlighted in \textbf{bold}.}
% \label{tab:model_evaluation_matrix}
% \begin{tabular}{l cccc c}
% \toprule
% \multirow{2}{*}{\textbf{Method}} & \multicolumn{4}{c}{\textbf{Seen Performance}} & \textbf{Unseen Generalization} \\
% \cmidrule(lr){2-5} \cmidrule(lr){6-6}
%  & PnP-Easy & PnP-Hard & Long-Horizon & \textbf{AVG} & \textbf{AVG} \\
% \midrule
% Model A         & 65.0 & 35.0 & 20.0 & 40.0 & 20.0 \\
% Model B         & 75.0 & 45.0 & 35.0 & 51.7 & 30.7 \\
% Model C         & 82.0 & 58.0 & 48.0 & 62.7 & 40.7 \\
% \textbf{JoyAI-RA (Ours)} & \textbf{95.0} & \textbf{85.0} & \textbf{78.0} & \textbf{86.0} & \textbf{76.3} \\
% \bottomrule
% \end{tabular}
% \end{table}

\subsection{Main Results}
\label{sec:main_results}

% \subsubsection{Robocasa}
% \label{sec:robocasa}
% [Placeholder] JoyAI-RA achieves state-of-the-art success rate on Robocasa across task categories. Full per-task numbers and analysis are provided in Table~\ref{tab:robocasa} (to be added).

% \subsubsection{Robotwin 2.0}
% \label{sec:robotwin}
% [Placeholder] On Robotwin~2.0, JoyAI-RA demonstrates strong performance under multi-embodiment and bimanual settings, indicating that the dual-alignment paradigm transfers across embodiments. Details in Table~\ref{tab:robotwin} (to be added).

\subsubsection{Real-World AgiBot Benchmark}
\label{sec:agibot}
\begin{figure}[t]
    \centering
    \includegraphics[width=1.0\columnwidth]{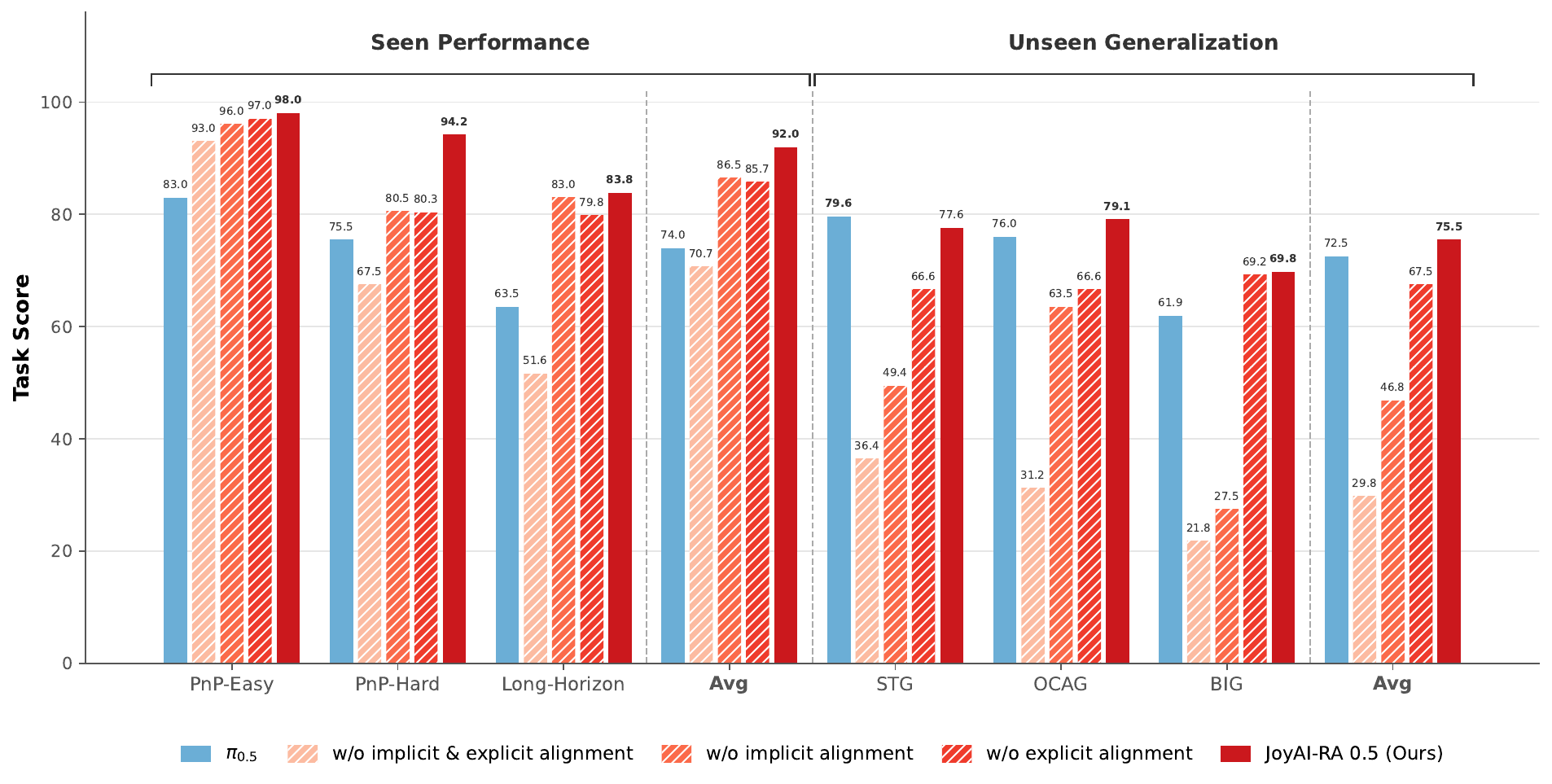}
    \caption{\textbf{Real-world task scores of different methods under seen performance and unseen generalization.} JoyAI-RA 0.5 achieves the highest average score in both settings. The alignment ablations further demonstrate the complementary contributions of implicit and explicit alignment to task execution and generalization.}
    \label{fig:main_results}
\end{figure}

Figure~\ref{fig:main_results} summarizes the results on seen tasks and unseen generalization on the Real-World AgiBot Benchmark, comparing JoyAI-RA 0.5 against the strong VLA baseline $\pi_{0.5}$~\cite{intelligence2025pi_} under an identical evaluation protocol.

For seen tasks, JoyAI-RA 0.5 clearly outperforms $\pi_{0.5}$, reaching an average task score of \textbf{92.0} against 74.0, and the advantage widens as task difficulty increases from pick-and-place to precise and long-horizon manipulation. This indicates that the physically grounded supervision from explicit alignment is especially beneficial where accurate contact and consistent long-range execution are required.

For unseen generalization, JoyAI-RA 0.5 achieves the highest average score, outperforming $\pi_{0.5}$ overall and across most dimensions. The largest gain appears under background and illumination variations, consistent with the visual and interaction diversity transferred from large-scale human egocentric videos through implicit alignment. JoyAI-RA 0.5 also performs best on novel objects and attributes. The only exception is spatial and topological generalization, where $\pi_{0.5}$ maintains a slight advantage, potentially reflecting the strong spatial priors of its vision-language backbone.

Overall, JoyAI-RA 0.5 achieves substantially stronger in-distribution performance while remaining competitive in unseen settings and demonstrating greater robustness to appearance shifts. These results demonstrate that the dual-alignment paradigm enables both reliable task execution and robust generalization.

% \begin{table}[htbp]
% \centering
% \footnotesize
% \setlength{\tabcolsep}{3.5pt}
% \caption{\textbf{Quantitative comparison of task scores between seen performance and unseen generalization.} Best results are highlighted in \textbf{bold}.}
% \label{tab:model_evaluation_matrix}
% \begin{tabular}{l cccc cccc}
% \toprule
% \multirow{2}{*}{\textbf{Method}} & \multicolumn{4}{c}{\textbf{Seen Performance}} & \multicolumn{4}{c}{\textbf{Unseen Generalization}} \\
% \cmidrule(lr){2-5} \cmidrule(lr){6-9}
%  & PnP-Easy & PnP-Hard & Long-Horizon & \textbf{AVG} & STG & OCAG & BIG & \textbf{AVG} \\
% \midrule
% $\pi_{0.5}$\cite{intelligence2025pi_}    & 83.0 & 75.5  & 63.5 & 74.0 & \textbf{79.6} & 76.0 & 61.9 & 72.5  \\
% % JoyAI-RA 0.1\cite{zhang2026joyai}     & 53.0 & 62.2 & 56.0 & 57.1  & 22.9 &29.1 &30.7 & 27.6 \\
% \textbf{JoyAI-RA 0.5(Ours)} & \textbf{98.0} & \textbf{94.2} & \textbf{83.8} & \textbf{92.0} & 77.6 & \textbf{79.1} & \textbf{69.8} & \textbf{75.5}  \\
% \bottomrule
% \end{tabular}
% \end{table}

% JoyAI-RA outperforms the strongest baseline by \textbf{+23.3} points in seen average and \textbf{+35.6} points in unseen generalization, confirming that the dual-alignment paradigm and the inner-outer RL stage jointly improve both task performance and generalization.

\subsubsection{RL Results}
\label{sec:rl_results}

To evaluate the effectiveness of the proposed inner-outer loop RL framework, we conduct real-world PnP experiments on mouse and headphones. It should be noted that in these experiments, the object placement regions are significantly expanded beyond the spatial distribution observed during SFT post-training, aiming to evaluate the policy generalization capability under unseen positions. For reference, on the mouse PnP task without positional distribution shift, the proposed RL method increases the success rate to 100\% within only a few training episodes; nevertheless, we focus primarily on the more challenging setting that requires generalization to unseen object positions.

\begin{figure}[t]
    \centering
    \includegraphics[width=0.6\columnwidth]{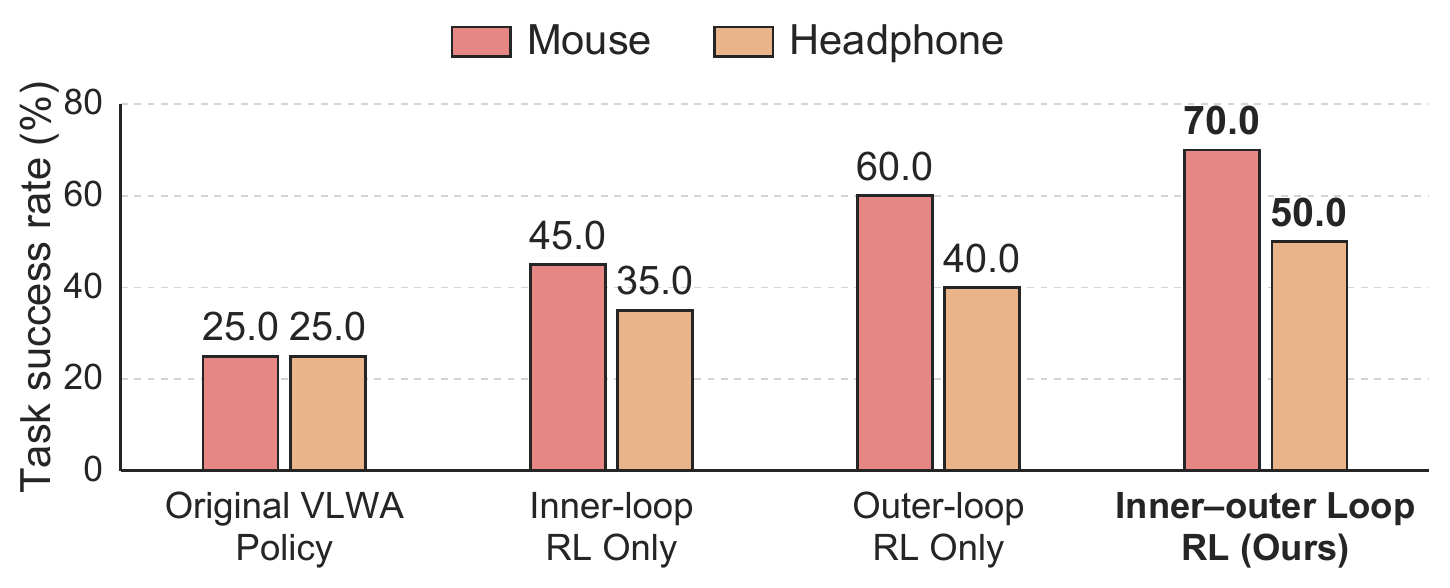}
    \caption{
    \textbf{Success-rate comparison of different RL strategies on pick-and-place tasks.}
    }
    \label{fig:rl_results}
\end{figure}

We compare four strategies under the same RL training episodes: the original VLWA policy, inner-loop-only adaptation, outer-loop-only adaptation, and the proposed inner-outer loop RL framework. As shown in Figure~\ref{fig:rl_results}, both the inner and outer loops improve performance over the original VLWA policy, while their combination achieves the best results on both tasks. The outer loop improves the generalization capability of the foundation VLWA by incorporating accumulated interaction data, whereas the inner loop enables rapid task-specific adaptation through residual policy optimization. These results demonstrate the complementary benefits of the two loops under positional distribution shifts.

% \begin{table}[htbp]
% \centering
% \footnotesize
% \setlength{\tabcolsep}{8pt}
% \caption{
% \textbf{Task score comparison of different RL strategies on pick-and-place tasks.}
% }
% \label{tab:rl_results}
% \begin{tabular}{lcc}
% \toprule
% \textbf{Method} & \textbf{Mouse} & \textbf{Headphone} \\
% \midrule
% Original VLWA Policy & 25.0 & 25.0 \\
% Inner-loop RL Only & 45.0 & 35.0 \\
% Outer-loop RL Only & 60.0 & 40.0 \\
% Inner-Outer Loop RL (Ours) & \textbf{70.0} & \textbf{50.0} \\
% \bottomrule
% \end{tabular}
% \end{table}

To maintain stable off-policy optimization, the current implementation synchronizes the updated VLWA weights from the outer loop at a relatively low frequency. More frequent synchronization can introduce substantial training instability, and developing a robust high-frequency synchronization mechanism remains an important direction for future work.

\label{sec:ablation}

\subsection{Ablation Studies}
We conduct ablation studies to assess the respective contributions of dual alignment (Sec.~\ref{sec:dual-alignment-ablation}), the world model (Sec.~\ref{sec:model-arch-ablation}), and latent-action conditioning (Sec.~\ref{sec:lac-ablation}). The dual-alignment ablation is evaluated on the full benchmark, whereas the other ablations use a fixed subset of PnP-Easy and PnP-Hard tasks in the desk scene for controlled comparison. We refer to the results on this subset under seen and unseen conditions as Desk (seen) and Desk (unseen), respectively.

\subsubsection{Dual Alignment Ablation Studies}\label{sec:dual-alignment-ablation}
We compare the complete JoyAI-RA 0.5 with three variants that remove one or both alignment mechanisms. In the variant without implicit alignment, Stage~1 is removed and the LAC-WM is replaced with an off-the-shelf open-source WM that has not undergone latent-action-conditioned pretraining. In the variant without explicit alignment, embodiment-specific action dimensions are directly concatenated into a shared vector, with unsupported dimensions zero-padded, bypassing the unified action representation—that is, neither mapping actions into the 130-dimensional canonical action space nor expressing end-effector motion as camera-frame chunk-relative end-effector actions—and thus without shared physical semantics. The third variant combines both modifications. All remaining training stages and evaluation settings are kept unchanged, and the results on the full benchmark are summarized in Figure~\ref{fig:main_results}. %Table~\ref{tab:model_evaluation_matrix_dual_alignment}.

The full model achieves the highest task scores under both seen and unseen conditions, reaching 92.0 and 75.5 on average. Removing both alignment mechanisms causes the largest drop, with the unseen average falling to 29.8, demonstrating that naively combining heterogeneous trajectories without transferable dynamics pretraining or semantic action alignment generalizes poorly. Removing implicit alignment mainly hurts unseen performance (46.8) while leaving seen performance relatively strong, with the degradation most pronounced under background and illumination variations—indicating that latent-action-conditioned video pretraining is critical for transferring the visual and interaction diversity of human videos. Removing explicit alignment produces a different failure pattern: unseen performance remains comparatively strong, but seen performance weakens to 85.7, most notably on precise PnP-Hard tasks, suggesting that concatenating and padding embodiment-specific actions without a unified action-space representation cannot provide the physically consistent supervision required for precise execution.

Overall, implicit alignment primarily strengthens transferable dynamics and unseen generalization, whereas explicit alignment improves physical grounding and execution precision. Their combination achieves the strongest performance across all seen tasks and all unseen dimensions, confirming that the two alignment channels provide distinct yet complementary supervision from heterogeneous data.

\subsubsection{Model Architecture Ablation Studies}
\label{sec:model-arch-ablation}

% In this subsection, we conduct a focused investigation into the role of the WM module within our proposed JoyAI-RA 0.5 framework. The core motivation is to assess whether introducing explicit future dynamics prediction—without any task-specific fine-tuning—can consistently benefit downstream decision-making across both physical and simulated environments. To this end, we adopt a decoupled and lightweight setup: we keep both the off-the-shelf VLM and the WM module frozen, and only leverage post-training trajectory data to evaluate their collaborative effect. This design choice intentionally isolates the contribution of the WM module from other confounding factors, such as representation fine-tuning or policy co-adaptation, thereby providing a cleaner ablation baseline. The quantitative results are summarized in Table~\ref{tab:wm_ablation}.

We conduct a controlled ablation study to determine whether incorporating the WM improves downstream control performance. We compare two otherwise identical JoyAI-RA post-training configurations, one without the WM and one with the WM. In both configurations, the off-the-shelf VLM is trainable, and the WM is frozen when included. Both configurations use the same post-training trajectory data, while all other components and training settings remain unchanged. Therefore, the performance difference directly reflects the contribution of the WM module. The quantitative results are summarized in Table~\ref{tab:wm_ablation}.

\begin{table}[htbp]
    \centering
    \caption{\textbf{Ablation results of the WM module in the JoyAI-RA 0.5 framework.}}
    \begin{tabular}{lccc}
        \toprule
        \textbf{Module} & \textbf{Desk (seen) Score} & \textbf{Desk (unseen) Score} & \textbf{AVG}\\
        \midrule
        w/o WM & 59.3 & 37.5  & 48.4 \\
        w/ WM & 62.3& 40.6  & \textbf{51.5} \\
        \bottomrule
    \end{tabular}
    \label{tab:wm_ablation}
\end{table}

As shown, the integration of the WM module yields consistent performance gains across all evaluation settings. On real-robot tasks, the task score improves from 59.3 to 62.3 on seen scenes, and more notably from 37.5 to 40.6 on unseen scenes, an absolute increase of 3.1 points in the more challenging generalization setting. This discrepancy between seen and unseen improvements suggests that the WM module contributes as a generalizable prior that becomes particularly valuable when visual or semantic distributions shift. 
% In the simulated Robotwin 2.0 benchmark, the advantage is even more pronounced, with the success rate climbing from 81.28\% to 89.22\%, yielding a relative improvement of nearly 10\%. The larger margin in simulation can be attributed to the more consistent and noise-free dynamics, which allow the WM’s predictive representations to align more faithfully with the ground-truth future states.

\subsubsection{Ablation on Latent-Action Conditioning for World-Model Training}\label{sec:lac-ablation}

To isolate the contribution of latent-action conditioning during world-model training, we compare a standard WM with the proposed LAC-WM under the same training and downstream evaluation setting. The two variants use the same world-model architecture, while only LAC-WM receives latent actions inferred from visual transitions as an additional conditioning signal during pretraining. During downstream policy training and evaluation, neither WM receives latent actions; the comparison therefore isolates whether latent-action-conditioned pretraining improves the dynamics representations transferred to the policy, rather than providing additional inference-time information. Table~\ref{tab:wm_latent_action_input_ablation} shows the results.

\begin{table}[htbp]
    \centering
    \caption{\textbf{Ablation results of latent-action conditioning during world-model training.}}
    \label{tab:wm_latent_action_input_ablation}
    \begin{tabular}{lcc}
        \toprule
        \textbf{World Model} & \textbf{Latent-Action Conditioning} & \textbf{Desk (seen) Score}  \\
        \midrule
        WM & No & 87.3 \\
        LAC-WM & Yes & \textbf{92.1} \\
        \bottomrule
    \end{tabular}
\end{table}

Latent-action conditioning improves the seen-task score from 87.3 to 92.1, an absolute gain of \textbf{4.8 points}. A standard WM must infer future evolution from the current observation and task context alone, even though multiple action-dependent futures may be plausible. In contrast, LAC-WM receives a compact transition-level condition during pretraining, which reduces this ambiguity and encourages the model to learn action-sensitive dynamics. The improvement supports using latent actions not only to incorporate action-unlabeled videos, but also to provide a more informative training signal for the world model.

\subsection{Human Ego-Video Data Scaling Analysis}
\label{sec:scaling}
% In this section, we investigate whether human egocentric video provides a scalable source of knowledge that can be transferred to downstream robot control. We first characterize the scale and diversity of EgoLive and then conduct two complementary studies. The first study primarily focuses on the diversity and scalability of the EgoLive dataset. By utilizing the LAM Encoder to annotate latent actions as corpus-wide surrogate action targets, we perform direct egocentric video pretraining, followed by identical robot-action post-training. Plain latent action supervision is adopted because it can be consistently obtained from all visual observation clips, it evaluates the diversity and scalability of EgoLive, and whether the increasing quantity of it used can be converted into performance improvements on robot policies. The second study varies the amount of human video (which includes EgoLive) used to pretrain the LAC-WM while keeping the robot data and downstream policy training fixed, thereby isolating the effect of data scale on dynamics learning as well as the VLWA framework. Together, these studies evaluate the benefits of scaling human egocentric video at both the self-constructed data level and the component level.

In this section, we investigate whether human egocentric video provides a scalable source of knowledge that can be transferred to downstream robot control. We first characterize the scale and diversity of our EgoLive dataset in Sec.~\ref{subsec:egolive}, followed by two complementary studies. 
% The first study focuses primarily on the diversity and scalability of EgoLive (Sec.~\ref{subsec:ego_scaling_effects}). 
The first study examines the scaling properties of EgoLive (Sec.~\ref{subsec:ego_scaling_effects}) by assessing how increasing the volume of egocentric pretraining data affects final robot policy performance, using a policy learning pipeline that maintains consistent supervision across all video clips. The second study isolates the effect of human video quantity on dynamics learning by varying the pretraining data composition for LAC-WM while keeping robot data and downstream training fixed (Sec.~\ref{subsec:lacwm_scaling}). Together, these two studies evaluate the benefits of scaling human egocentric video at both the dataset level and the component level.

\begin{figure}[h]
    \centering
    \includegraphics[width=0.9\linewidth]{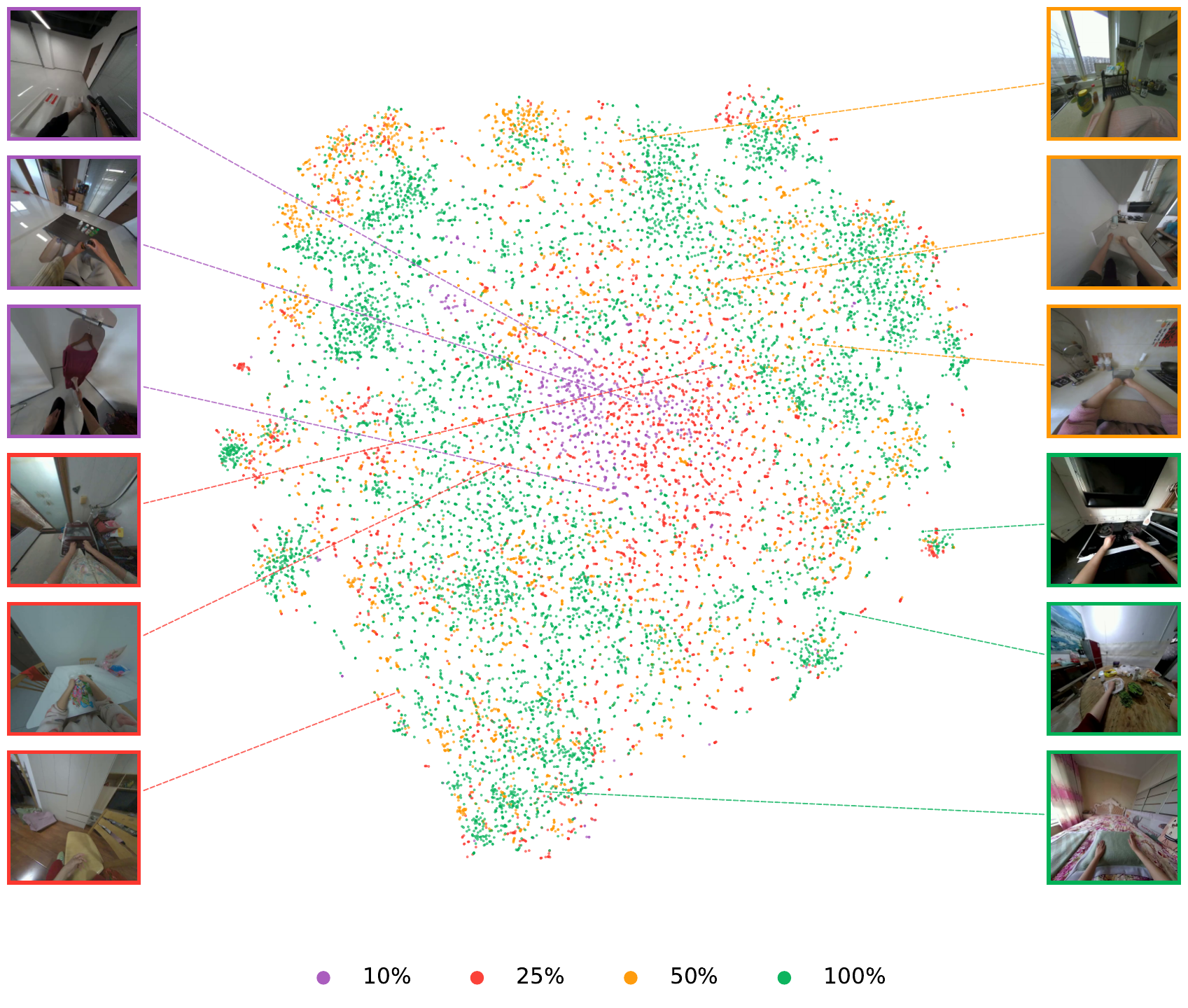}
    \caption{\textbf{T-SNE visualization of the action-embedding distributions of EgoLive at different dataset scales (10\%, 25\%, 50\%, and 100\%).} The four data fractions form nested subsets, such that each smaller-scale dataset is fully contained in all larger-scale datasets.}
    \label{fig:egolive_distribution}
\end{figure}

\subsubsection{Scale and Diversity of EgoLive}
\label{subsec:egolive}

EgoLive captures a diverse spectrum of task contexts, action behaviors, and object interactions. Originating from broad real-world settings, the dataset contains a massive volume that spans more than 20,000 demonstration hours. Further details regarding the EgoLive dataset are elaborated in the Appendix \ref{subsec:egolive_dd}.

As shown in Figure~\ref{fig:egolive_distribution}, we use an action-centric representation model to extract video-level embeddings and jointly apply t-SNE to samples from four dataset scales—10\%, 25\%, 50\%, and 100\%—projecting them onto the same two-dimensional plane for direct comparison. The representation primarily captures action patterns, interaction processes, and behavioral variations rather than relying solely on scene or object appearance. As the dataset grows, the projected samples cover a broader area and extend into more peripheral and low-density regions. This suggests that additional data not only increases the density of existing action patterns but also introduces new action types, execution styles, and interaction modes, thereby improving coverage and diversity in the underlying action representation space.

\subsubsection{Scaling with EgoLive Dataset}
\label{subsec:ego_scaling_effects}
Following the data distribution analysis of EgoLive, we next investigate whether the benefits of scaling the self-constructed dataset can extend to policy pretraining and transfer to downstream robot control. We progressively scale the EgoLive pretraining data using nested subsets containing 10\%, 25\%, 50\%, and 100\% of the full corpus of over 20,000 hours, and evaluate the resulting models on eight downstream tasks from the PnP-Easy and PnP-Hard benchmarks.

% All experiments are conducted with pre-training on an unfinetuned VLM and a WM, followed by direct post-training. A key challenge in utilizing public human video datasets is the pervasive lack of consistent hand pose annotations, many of which are either missing or noisy, thereby preventing supervision through explicit hand poses from scaling across the entire corpus. To overcome this limitation, we adopt latent actions, which learn a shared action vocabulary directly from raw videos, and accordingly employ the resulting latent actions as the supervision targets during pre-training.
To isolate the impact of ego-video scaling, we adopt a pre-training configuration where the volume of the EgoLive dataset is the sole variable. The model is initialized from an open-source, pre-trained VLM and WM. Specifically, the world model remains frozen, while the VLM and action expert models are fully trained on the EgoLive dataset. Upon completing the egocentric pre-training, all models undergo an identical robotic post-training pipeline. This analysis complements the subsequent LAC-WM scaling experiment by testing whether the scaling benefit holds under the EgoLive dataset.
% To isolate the effect of the ego-video scale, we adopt a controlled surrogate-action pretraining setting. The model is initialized from an open-source pretrained VLM and WM. The WM remains frozen, while the VLM and Action Expert are trained exclusively on EgoLive using latent actions inferred from visual transitions as surrogate action targets. After ego pretraining, all models undergo the same robot post-training procedure. 
% This design is important because reliable hand-pose annotations cover only a subset of the corpus and vary in quality. Directly scaling hand-pose supervision would therefore conflate video quantity with annotation coverage and accuracy. In contrast, l
% Latent actions provide consistent supervision for all visually valid clips, allowing us to measure whether additional ego video yields a more transferable policy initialization. This analysis complements the subsequent LAC-WM experiment by testing whether the scaling benefit holds under the EgoLive dataset.
% This policy-level analysis complements the preceding LAC-WM experiment by testing whether the scaling benefit persists beyond dynamics-model pretraining.

\begin{figure}[ht]
    \centering
    \includegraphics[width=0.8\linewidth]{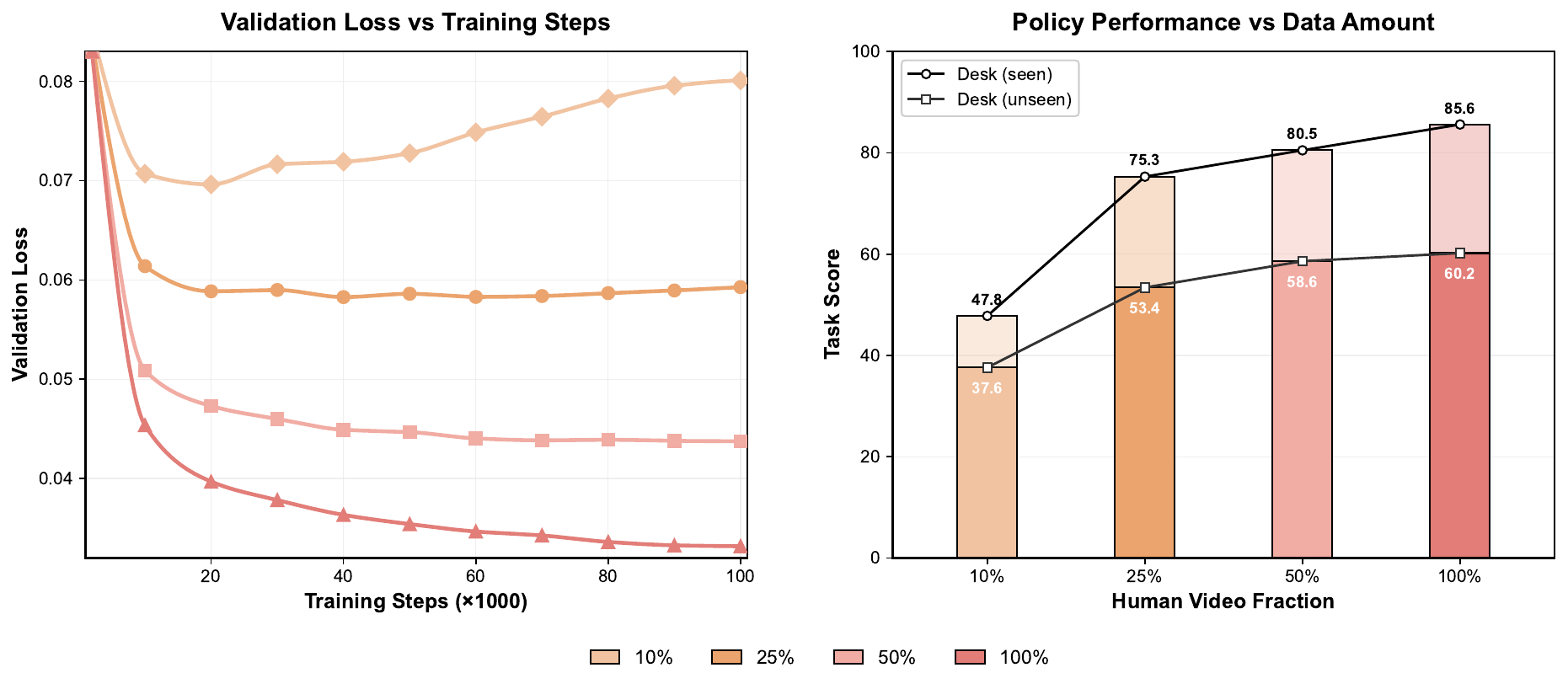}
    \caption{\textbf{Scaling with EgoLive dataset.} Left: held-out human validation loss over pretraining steps using 10\%, 25\%, 50\%, and 100\% of EgoLive. Right: seen- and unseen-task task scores after identical robot-action post-training. Higher, lighter bars denote seen tasks, while lower, saturated bars denote unseen tasks.}
    \label{fig:retarget5_pct_val_success}
\end{figure}

% The results shown on the left of Fig.~\ref{fig:retarget5_pct_val_success} demonstrate that increasing the scale of pre-training ego video data reduces the held-out validation loss. The relationship between training steps and human validation loss across varying data scales: with smaller datasets (10\%), the validation loss declines initially but soon flattens or even rises, a clear symptom of overfitting due to limited behavioral coverage. By contrast, larger datasets (50\%-100\%) enable sustained and stable loss reduction throughout the entire training process, with no overfitting observed. This suggests that scaling up ego-video data enriches the diversity of visual and behavior patterns in the input and output space, thereby continuously strengthening the model's cross-modal representation learning and preventing premature convergence.

As shown on the left of Fig.~\ref{fig:retarget5_pct_val_success}, increasing the amount of ego-video pretraining data consistently improves held-out validation performance. With only 10\% of the corpus, the validation loss initially decreases but subsequently rises, indicating overfitting under limited behavioral coverage. In contrast, the 50\% and 100\% settings maintain stable loss reduction throughout pretraining, with the full dataset achieving the lowest validation loss. These results indicate that larger ego-video corpora provide more diverse visual transitions and manipulation behaviors, improving the generalization of  policy pretraining.

% This scaling effect is also evident in downstream performance, with improvements in both in-domain success rates and out-of-domain generalization, as shown on the right of Fig.~\ref{fig:retarget5_pct_val_success}. As the data volume increases from 10\% to 100\%, the in-domain success rate rises from 47.8\% to 85.6\%, while the out-of-domain success rate improves from 37.6\% to 60.2\%. The performance curve shows no tendency to plateau within the tested range. This strongly suggests that, despite the inherent noise, lack of constraints, and task misalignment in human egoview demonstrations, scaling up human data consistently enriches the prior knowledge available for manipulation tasks.

The benefit transfers consistently to downstream robot control, as shown on the right of Fig.~\ref{fig:retarget5_pct_val_success}. As the ego-video fraction increases from 10\% to 100\%, the seen-task score improves from 47.8 to 85.6, while the unseen-task score rises from 37.6 to 60.2. Both metrics improve across every tested data scale, showing that the gains acquired from human video remain effective.
% after the supervision target is switched from latent actions to physical robot actions. 
The present experiment adopts a more challenging pure-ego setting to show that increasing human video can also produce progressively more transferable policy initialization before physical-action grounding.

Overall, the empirical results underscore the robust scalability of the EgoLive dataset. On one hand, scaling up data from 10\% to the full dataset consistently enhances pretraining generalization and resolves limited-coverage overfitting. On the other hand, the benefits acquired from scaling up human video seamlessly transfer to downstream physical robot control, achieving steady performance boosts in both seen and unseen environments across all tested data scales.

% Together with the LAC-WM scaling results, this experiment demonstrates that increasing human egocentric video strengthens both dynamics learning and transferable policy initialization, establishing its broader system-level value for scalable robot learning.

% It is worth noting that the absolute task scores in this study are not expected to match those of the preceding LAC-WM scaling study in Section~\ref{subsec:lacwm_scaling}, as the two experiments examine complementary uses of human video under different training paradigms. The former evaluates how data scale improves latent-action-conditioned dynamics learning, while the present experiment adopts a more challenging pure-ego setting to show that increasing human video can also produce progressively more transferable policy initialization before physical-action grounding.

\subsubsection{Scaling LAC-WM with Human Ego-Video}
\label{subsec:lacwm_scaling}

Following the dataset-level analysis of EgoLive, we next investigate whether scaling human video produces transferable gains through the LAC-WM, the dynamics component that directly learns from heterogeneous video data. Our full human-video corpus contains approximately 53K hours, and the data fractions below are measured relative to this corpus. 

\begin{wrapfigure}{r}{0.44\textwidth}
    \vspace{-10pt}
    \centering
    \includegraphics[width=\linewidth]{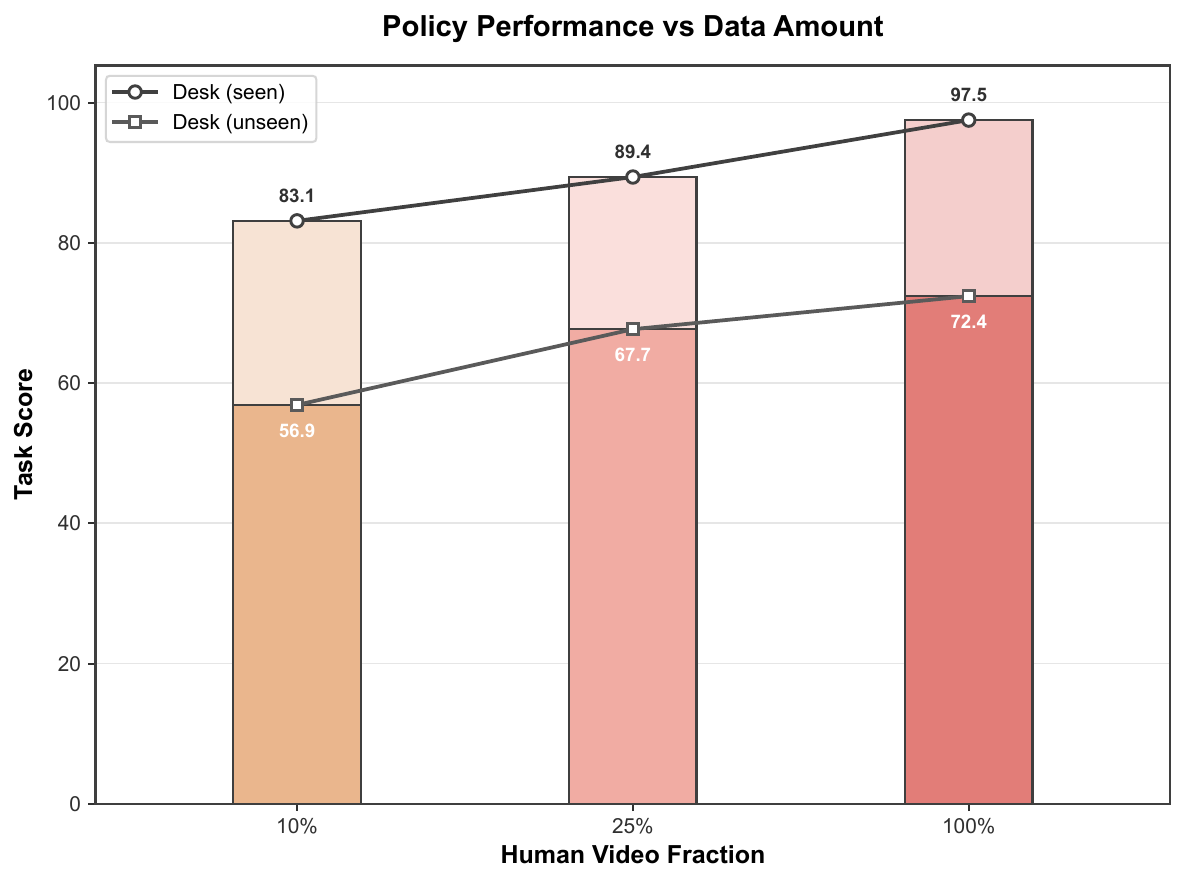}
    \captionsetup{
        font=small,
        justification=raggedright,
        singlelinecheck=false
    }
    \caption{
        \textbf{Scaling LAC-WM with human ego-video.}
        Seen- and unseen-task performance improves monotonically as the
        human-video fraction increases, while the robot data and downstream
        protocol remains fixed, demonstrating a stronger transferable
        dynamics prior.
    }
    \label{fig:lacwm_scaling}
    \vspace{-10pt}
\end{wrapfigure}
Across all settings, we hold the robot trajectory corpus fixed and vary only the amount of human ego video. We apply the same latent-action labeling and LAC-WM pretraining pipeline in each setting, freeze the resulting LAC-WM, and incorporate it into an identical downstream policy training and evaluation protocol. This controlled component-level study isolates the contribution of additional human visual experience and tests whether a larger human-video corpus can be converted into a more effective dynamics prior for downstream control. We report task scores on the standard seen-task test and unseen-task generalization test in Figure~\ref{fig:lacwm_scaling}.
% \begin{table}[h!]
%     \centering
%     \caption{
%     \textbf{Scaling LAC-WM pretraining with increasing fractions of the human-video corpus while keeping the robot dataset and downstream protocol fixed.} Fractions are rounded to match the percentage-based convention used in the policy-level scaling study.
%     }
%     \label{tab:lacwm_scaling}
%     \begin{tabular}{lccc}
%         \toprule
%         \textbf{Human Video Fraction}
%         &  \textbf{Desk (seen) Score} & \textbf{Desk (unseen) Score} & \textbf{AVG}\\
%         \midrule
%         10\% & 83.13 & 56.88 & 70.01 \\
%         25\% & 89.38 & 67.68 & 78.53\\
%         100\% & 97.50 & 72.40 & \textbf{84.95}\\
%         \bottomrule
%     \end{tabular}
% \end{table}

Increasing the human-video fraction from approximately 10\% to 25\% raises the seen-task score from 83.1 to 89.4, an improvement of \textbf{6.3 points}. The unseen-task score increases more substantially from 56.9 to 67.7, yielding a gain of \textbf{10.8 points}. Scaling to the full corpus further improves seen and unseen task scores to 97.5 and 72.4, respectively. The larger improvement on unseen tasks at the earlier scaling stage indicates that additional human video contributes more than in-distribution task familiarity and helps the LAC-WM acquire dynamics knowledge that transfers across task variations. Overall, the monotonic gains across data fractions show that, under fixed robot supervision and downstream training, expanding human visual experience produces a stronger dynamics prior and improves downstream robustness.

\WFclear

Together with the EgoLive scaling results, this experiment demonstrates that increasing human egocentric video strengthens both dynamics learning and transferable policy initialization, establishing its broader system-level value for scalable robot learning.

\section{Conclusion}
\label{sec:conclusion}

We propose JoyAI-RA~0.5, a Vision-Language-World-Action framework that addresses the central obstacle in scaling generalist manipulation: turning heterogeneous, weakly labeled data into a shared and transferable form of supervision. Architecturally, it combines a Latent-Action-Conditioned World Model and a Vision-Language Model to learn physical world-dynamics priors and visual-semantic understanding, respectively. Complementing this design, its dual-alignment paradigm routes each source to the supervision it can reliably provide—latent-action supervision for action-free videos and a unified action space for reliable trajectories—so that human egocentric video, simulation, and real-robot data reinforce rather than interfere with one another.

Our experiments substantiate this design along three fronts. On the Real-World AgiBot Benchmark, JoyAI-RA 0.5 clearly outperforms a strong VLA baseline on seen tasks, with the margin widening as manipulation becomes more precise and long-horizon, and remains competitive under unseen generalization while proving notably more robust to background and illumination shifts. Our ablations confirm that the two alignment channels are complementary rather than redundant: implicit alignment supplies transferable dynamics knowledge that underpins robustness to appearance and environment shifts, whereas explicit alignment provides the physical grounding that precise execution demands. Most importantly, performance improves consistently as human egocentric pretraining data increase, with no sign of plateauing at our largest scale. Taken together, these findings establish human egocentric video not merely as auxiliary data, but as a primary axis along which real-world manipulation capability can be scaled.

\bibliographystyle{plainnat}
\bibliography{paper}

\newpage

\appendix

\section*{Appendix}

\section{Contributions}
\label{sec:contributions}

% \noindent\textit{Authors are listed in alphabetical order by last name; the ordering does not indicate relative contribution.}

% \vspace{0.8em}

% JoyAI-RA Team
\noindent
\begin{minipage}[t]{0.56\linewidth}
\raggedright
\textbf{Core Contributors}
\\[2pt]
{\small\textit{\textsuperscript{*}denotes Co-first authors, listed alphabetically by last name.}}
\begin{itemize}[
    label={},
    leftmargin=0pt,
    itemindent=0pt,
    labelsep=0pt,
    itemsep=2pt,
    topsep=4pt
]
    \item Dafeng Chi\textsuperscript{*}
    \item Peidong Liu\textsuperscript{*}
    \item Zhiyuan Xiang\textsuperscript{*}
    \item Sheng Xu\textsuperscript{*}
    \item Tianle Zhang\textsuperscript{*}
    \item Yuzheng Zhuang\textsuperscript{\dagger}
    \item Dongwei Li
    \item Bin Wang
    \item Zhihao Yuan
    \item Bowen Yang
    \item Mingyang Li
    \item Wenhao Li
    \item Linbo Zhai
    \item Junjie Wang
    \item Jiawei Li
    \item Ao Li
    \item Chenyu Wu
    \item Yihang Li
    \item Shibo Jin
    \item Daming Wang
    \item Kangliang Chen
    \item Nan Duan
    \item Liang Lin\textsuperscript{\dagger}
\end{itemize}
\end{minipage}
\hfill
\begin{minipage}[t]{0.42\linewidth}
\raggedright
\textbf{Contributors}
\begin{itemize}[
    label={},
    leftmargin=0pt,
    itemindent=0pt,
    labelsep=0pt,
    itemsep=2pt,
    topsep=4pt
]
    \item Yunchen Cai
    \item Peng Cao
    \item Zengjue Chen
    \item Yuqi Cheng
    \item Tianchen Deng
    \item Yicheng Gong
    \item Chenguang Gui
    \item Qingrong He
    \item Ruodai Li
    \item Jiaming Liang
    \item Xiangkai Ma
    \item Junnan Nie
    \item Xing Pan
    \item Hui Shen
    \item Jiahao Sun
    \item Hanwen Wan
    \item Song Wang
    \item Huangtao Wu
    \item Junwu Xiong
    \item Junzhe Xiong
    \item Hang Xu
    \item Yifei Yu
    \item Xin Yi
    \item Chubin Zhang
    \item He Zhang
    \item Likui Zhang
\end{itemize}
\end{minipage}

\renewcommand{\thefootnote}{\fnsymbol{footnote}}
% \footnotetext[1]{Equal contribution, listed alphabetically by last name; the ordering does not indicate relative contribution.}
\footnotetext[2]{Corresponding authors: Yuzheng Zhuang \textless zhuangyuzheng.1@jd.com\textgreater, Liang Lin \textless linliang@ieee.org\textgreater.}
\renewcommand{\thefootnote}{\arabic{footnote}}

\section{Data Distribution of EgoLive}
\label{subsec:egolive_dd}

\begin{figure}[t]
    \centering
    \includegraphics[width=0.9\linewidth]{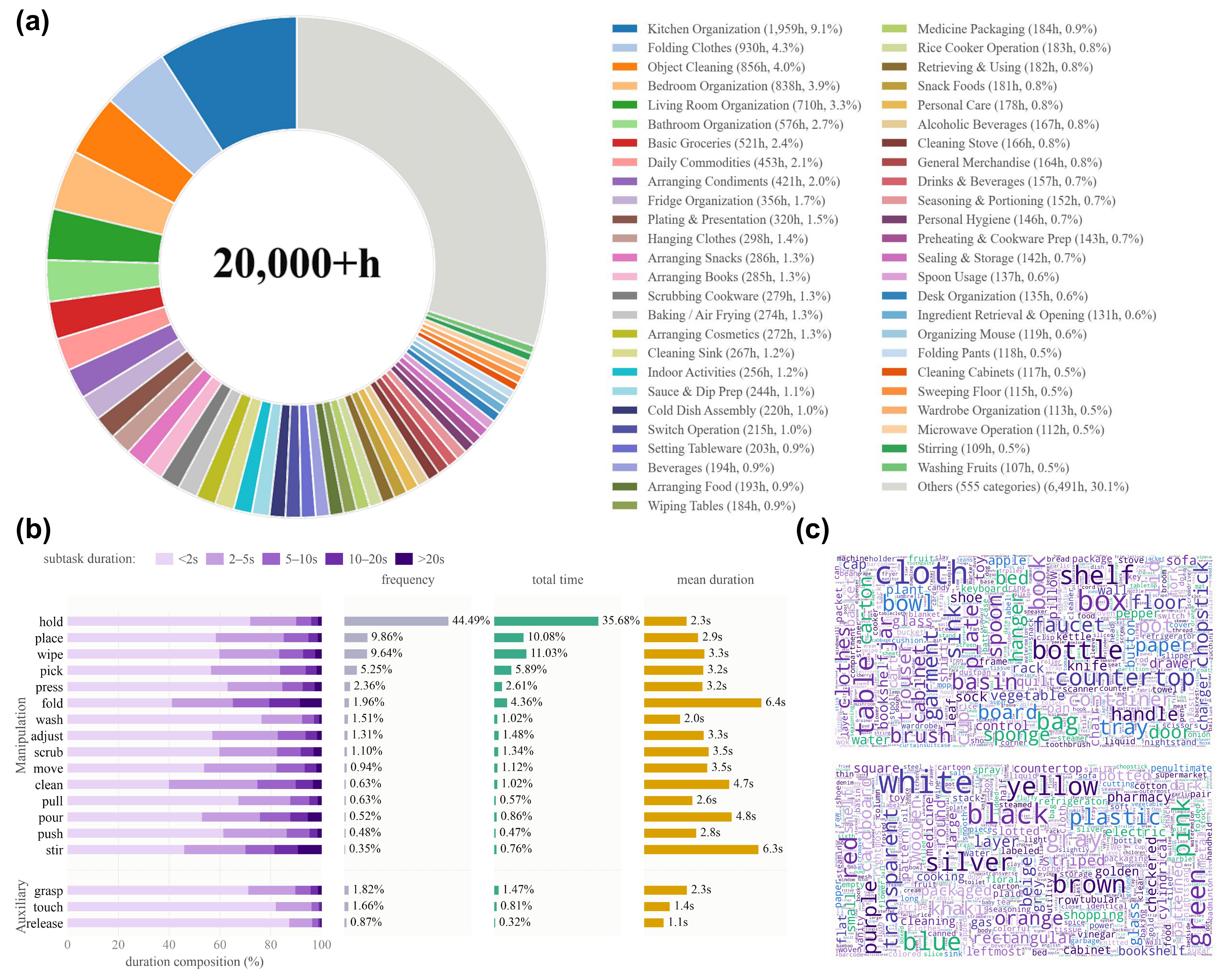}
    \caption{\textbf{Data Distribution of EgoLive.}
    EgoLive comprises 20,000+ hours of demonstrations, approximately 898,000 episodes, and 23.6 million subtasks.
    (a) Distribution of demonstration hours across more than 600 fine-grained task scenarios, highlighting the 50 most frequent categories.
    (b) Statistics for 18 core manipulation and auxiliary action types, including duration composition across five temporal bins, frequency, total duration, and mean duration.
    (c) Word clouds illustrating semantic diversity across 53,400 manipulated-object categories and 31,400 object-attribute categories, where font size indicates frequency across subtasks.}
    \label{fig:egolive-dataset}
\end{figure}

Figure~\ref{fig:egolive-dataset} illustrates the scale and diversity of EgoLive in terms of task scenarios, action patterns, manipulated objects, and object attributes, which comprises 20,000+ hours of demonstrations, approximately 898,000 episodes, and 23.6 million subtasks collected from a broad range of real-world activities. As shown in Figure~\ref{fig:egolive-dataset}(a), the dataset covers more than 600 fine-grained task scenarios, including kitchen organization, clothes folding, object cleaning, bedroom and living-room organization, food preparation, and everyday item manipulation. The distribution exhibits a pronounced long tail: kitchen organization, the largest individual category, accounts for 9.1\% of the total duration, whereas the 555 categories grouped as ``Others'' collectively contribute 30.1\% of the dataset. This broad coverage exposes models to both frequently occurring activities and diverse long-tail scenarios involving multiple objects, multiple steps, and extended task sequences.

Beyond scenario diversity, EgoLive provides fine-grained annotations along three complementary semantic dimensions: actions, manipulated objects, and object attributes. The complete annotation vocabulary contains 2,662 fine-grained action categories, and Figure~\ref{fig:egolive-dataset}(b) summarizes the temporal and occurrence statistics of 18 core action types, comprising 15 manipulation actions and three auxiliary actions. Their duration distributions across five temporal bins reflect the distinct temporal characteristics of actions such as holding, wiping, placing, folding, grasping, touching, and releasing. The substantial variation in both action frequency and total duration further highlights the diversity of action patterns represented in the dataset. As shown in Figure~\ref{fig:egolive-dataset}(c), the dataset further contains 53,400 manipulated-object categories and 31,400 object-attribute categories, covering diverse object identities and properties such as material, color, state, shape, and spatial configuration. These action-object-attribute combinations provide dense coverage of common manipulation patterns while retaining broad long-tail semantic diversity, supporting compositional generalization across previously unseen objects, attributes, and tasks.

\end{document}